\documentclass{article}
\usepackage{iclr2027_conference,times}
\iclrfinalcopy
\usepackage{natbib}
\setcitestyle{authoryear,round,citesep={;},aysep={,},yysep={;}}

\usepackage{amsmath,amsfonts,bm}

\def\eqref#1{equation~\ref{#1}}

\def\1{\bm{1}}

\DeclareMathAlphabet{\mathsfit}{\encodingdefault}{\sfdefault}{m}{sl}
\SetMathAlphabet{\mathsfit}{bold}{\encodingdefault}{\sfdefault}{bx}{n}

\usepackage{hyperref}
\hypersetup{hidelinks}
\usepackage{url}

\usepackage{booktabs}

\usepackage{graphicx}
\usepackage{float}
\usepackage{subcaption}
\usepackage{multirow}

\usepackage{amsmath,amssymb}
\usepackage{algorithm}
\usepackage{algpseudocode}

\usepackage{caption}

\usepackage{xcolor}

\definecolor{mainbar}{RGB}{85, 107, 47}
\definecolor{overallbar}{RGB}{60, 80, 100}
\definecolor{gridgray}{RGB}{230, 230, 230}

\usepackage{xcolor}

\usepackage{meta_arxiv}

\title{META-Robot: Self Cognition and Capability Scaling in Agentic Robots}
\author{Xuanzhi Feng\textsuperscript{1,*}, Zeyu Liu\textsuperscript{1,*}, Weiheng Gao\textsuperscript{1,*}, Xinchen Li\textsuperscript{1,*}\\
Jiankun Tang\textsuperscript{1}, Fangqi Zhu\textsuperscript{1}, Quanxin Shou\textsuperscript{1}, Hao Chen\textsuperscript{1}\\
Zhengyang Yan\textsuperscript{1}, Hualei Zhang\textsuperscript{1}, Junhao Li\textsuperscript{1}, Zhiyao Wu\textsuperscript{2}\\
Jie Zhang\textsuperscript{1}, Song Guo\textsuperscript{1,\ensuremath{\dagger}}\\[0.5em]
{\small\normalfont\textsuperscript{1}The Hong Kong University of Science and Technology}\\
{\small\normalfont\textsuperscript{2}University of Macau}\\[0.35em]
{\small\normalfont\textsuperscript{*}Equal contribution.\quad\textsuperscript{\ensuremath{\dagger}}Corresponding author.}}
\date{}
\begin{document}

\maketitle

\begin{abstract}
LLM-based robotic agent systems can generate programs for robot control, yet their generalization to new tasks often depends on elaborate scaffolding. Expanding their skill repertoires alone offers limited guidance on how to assess new capabilities or identify what remains beyond their reach. We introduce META-Robot, a framework that couples self cognition with capability scaling by expanding the criteria through which an agent evaluates and develops its skills. At its core, action-level predicates express explicit conditions for assessing execution and provide a structured basis for LLM reasoning. Using the resulting assessments, the agent identifies capability gaps, extends relevant evaluation criteria, and guides the structured expansion of its skills. Proprioceptive feedback further grounds this self-scaling process in the robot's physical state. Experiments in simulation and on real robots demonstrate improved overall task performance, generalization to new tasks, continued expansion of skills and evaluation criteria, and effective use of non-visual proprioceptive signals.
\end{abstract}

\section{Introduction}
 Robotic agents have emerged as a promising paradigm for general-purpose robotic manipulation recently with generated code as control policy \cite{liang2023code, xiao2026enpire}. While  these systems \cite{kumar2026act} can adapt to new tasks without training to some extent, their scalability remains limited. When tasks require evaluation criteria, feedback modalities, or skills beyond predefined scaffolds, agents cannot revise these system-level limitations, leading to repeated strategy switching and compounding ineffective modifications that make meta-level task generalization or reliable recovery nearly impossible.

Predicate scaling offers a natural solution by allowing agents to retain and reuse predicates acquired from previous tasks. This accumulation can improve transfer and reduce the need for additional predefined scaffolding. However, scaling predicates alone cannot overcome fixed system scaffolding, as systems cannot have explicit cognition of task and its own capability limits. Recent efforts to improve scalability follow a primary strategy: replacing accumulation of fixed predicates with validated, agent-discovered skills distilled from execution processes~\cite{zhang2026playful,lu2026aspire}. This reduces manual engineering and allows past repairs to guide future tasks. However, these skills are accumulated without structured links to their  expected effects and required evidence. Consequently, the skill library may continue to grow while the system's understanding of the task and its own capabilities remains unchanged, limiting substantive capability scaling.

The emergence of Neuro-Symbolic Predicates (NSPs)~\cite{liang2025visualpredicator,athalye2026pixels} presents a compelling foundation for scaling robot  agent system cognition beyond fixed mechanisms. By combining neural perceptual queries with programmatic logic, NSPs can construct abstractions for evaluation, offering  a path toward adapting to new evaluation criteria. Nevertheless, extending predicate invention to self cognition and capability scaling in robotic agent systems remains challenging. Predicates must assess not only whether task goals are satisfied, but also whether the current system has sufficient observations, evaluation criteria, and executable skills. We argue that grounding predicates with physical evidence at the action level is crucial, as it allows the agent to exploit available proprioceptive signals when vision is insufficient, construct corresponding skills and evaluation criteria, and integrate them into control and verification, effectively coupling cognitive and capability scaling with evidence-grounded assessment.

\begin{figure}[!htbp]
    \centering
    \includegraphics[width=1\linewidth]{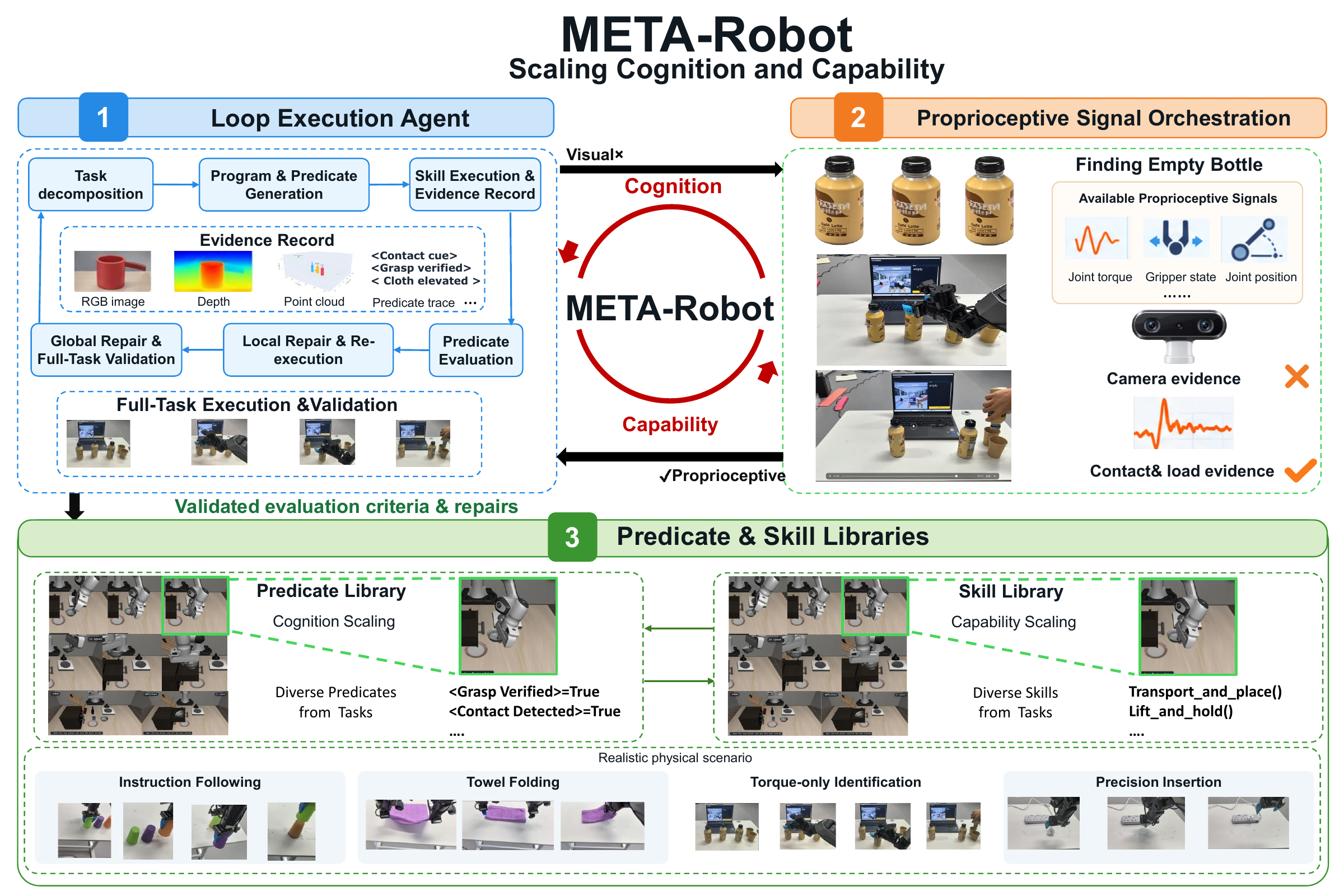}
    \caption{META-Robot starts from a task description. Its framework consists of three components: (1) Loop Execution Agent, which generates programs and predicates, evaluates execution evidence, and repairs failures; (2) Proprioceptive Signal Orchestration, which supplements insufficient visual observations with physical feedback; and (3) coupled Predicate and Skill Libraries, which consolidate validated criteria and repairs for cognitive and capability scaling.}
    \label{fig:overview}
\end{figure}

To this end, we propose META-Robot, as illustrated in Fig.~\ref{fig:overview}, a system for self cognition and capability scaling that grounds robotic meta-generalization in evidence-based, action-level predicates for both system self-assessment and task evaluation. By recording predicate transitions around skill executions, coupling validated predicates with reusable skills, and orchestrating proprioceptive feedback, META-Robot creates a scalable system for jointly scaling  robotic cognition and capabilities.

META-Robot incorporates several key innovations. First, to mitigate the ambiguity between coarse execution logs and actual task-state changes, we introduce a closed-loop execution agent that evaluates each action execution through action-level predicate transitions. The agent records observations, inputs, outputs, and pre- and post-execution predicate states, associating program behavior with changes in robot action. When an error is detected, it performs targeted local repair and re-execution, followed by full-task validation of the revised program. Second, fixed evaluation criteria make failures beyond the coverage of the existing system difficult to diagnose and can lead to unstructured skill accumulation. To address this, META-Robot maintains coupled predicate and skill libraries that expand through validated execution experience. Although this design supports continual knowledge scaling, it also requires precise alignment between the two libraries. META-Robot establishes explicit predicate--skill associations to support grounded skill retrieval, composition, and expansion.

We conduct extensive experiments in both simulation environments [23] and real-world environments
to validate the effectiveness of META-Robot. (i) META-Robot achieves stronger overall performance than prior methods, reaching a success rate of 95.8\% on the LIBERO-Pro Spatial-Task, compared with 60.0\% for other baselines. Crucially, we observe (ii) emergent cognition and capabilities including the scaling of reusable skills and their corresponding evaluation criteria from validated execution experience. We further demonstrate (iii) robust generalization through zero-shot transfer to held-out long-horizon tasks without additional debugging and successful deployment across diverse real-robot tasks. Together with the ablation results, these findings establish META-Robot as a scalable framework for jointly expanding robotic cognition and capability. This paper makes the following contributions:

\begin{itemize}
    \item We introduce META-Robot, which scales its cognition and capability by jointly improving task policies and the supporting system itself through predicate-guided self-assessment.

    \item We develop coupled action-level predicate and skill libraries that support continual refinement of evaluation criteria and policies, as well as the generation of new structured criteria and skills beyond the existing repertoire.

    \item We propose a proprioceptive signal orchestration mechanism that expands the system's capacity for perception, evaluation, and control.
\end{itemize}
\section{Related Work}

\textbf{Agentic Robotic Systems.}
Recent work increasingly formulates robot control as a closed-loop agentic process rather than a one-shot mapping from language to actions. Systems such as Hi Robot and Being-0 combine high-level vision-language reasoning with lower-level VLA policies or modular skills~\cite{shi2025hi,yuan2025being}. Robot-use agents further expose capabilities through executable interfaces and stateful harnesses: CaP-X systematically studies Code-as-Policy agents across different abstraction levels and introduces structured execution feedback and visual differencing; Guava introduces semantic tools and multimodal observations; Thea adds persistent scene context and execution evaluation; and Show-Harness enables VLM-driven manipulation through semantic motion primitives, like Cap-X, Guava~\cite{fu2026cap,liu2026guava}. These works establish perception, planning, execution, verification, and recovery as core components of agentic robotics. However, most still rely on predefined tools, observation channels, and evaluators provided by the surrounding harness.

\textbf{Skill-Scaling Robotic Agents.}
Beyond closed-loop execution, skill-scaling agents retain and adapt experience across tasks. One line of work reuses prior programs or reflections as memory, including Memory Transfer Planning, Growing with Your Embodied Agent, DAHLIA, and EEAgent~\cite{kagaya2025memorytransfer,meng2025growing,meng2025embodied,wang2026eeagent}. Another expands the executable repertoire itself: Playful discovers reusable skills, ASPIRE distills validated repairs, RHO searches repository-level policies, and Embodied Skills, ALRM, and AOR broaden reusable tools or controller implementations~\cite{zhang2026playful,lu2026aspire,elmaaroufi2026rho,wang2026embodiedskills,santos2026alrm,kumar2026act}. Yet capability growth is typically decoupled from the evolution of evaluation criteria and evidence requirements. META-Robot instead couples hierarchical skills with action-level predicates and additional feedback channels, allowing capabilities and the mechanisms used to assess them to scale together.

\section{Method}
We introduce META-Robot, a novel framework for meta-level generalization to new tasks. The META-Robot framework is grounded in the scaling of cognition and capability, where evaluation criteria, feedback acquisition, available signals, and executable skills are jointly expanded, thereby reducing dependence on predefined scaffolding. META-Robot operates directly on  action-level predicates throughout the process, instead of coarse execution outcomes, which better connects program behavior with changes in task conditions. Specifically, the closed-loop execution agent records observations, inputs, outputs, and pre- and post-execution predicate states for every skill invocation and subtask end. The predicate-state trace enables the system to consistently assess progress, attribute failures, and verify targeted repairs through local re-execution before full-task validation. We also introduce coupled action-level predicate and Skill Libraries and a proprioceptive signal orchestration mechanism to overcome the limitations of unstructured skill accumulation and vision-only verification. With these components, META-Robot can enrich its task evidence and cognition, consolidate validated repairs as reusable skills, and acquire additional physical evidence when existing observations are insufficient, progressively expanding both system cognition and capability. Fig.~2 illustrates an overview of the META-Robot framework.

\subsection{loop execution agent system}
\begin{figure}[!htbp]
    \centering
    \includegraphics[width=1\linewidth]{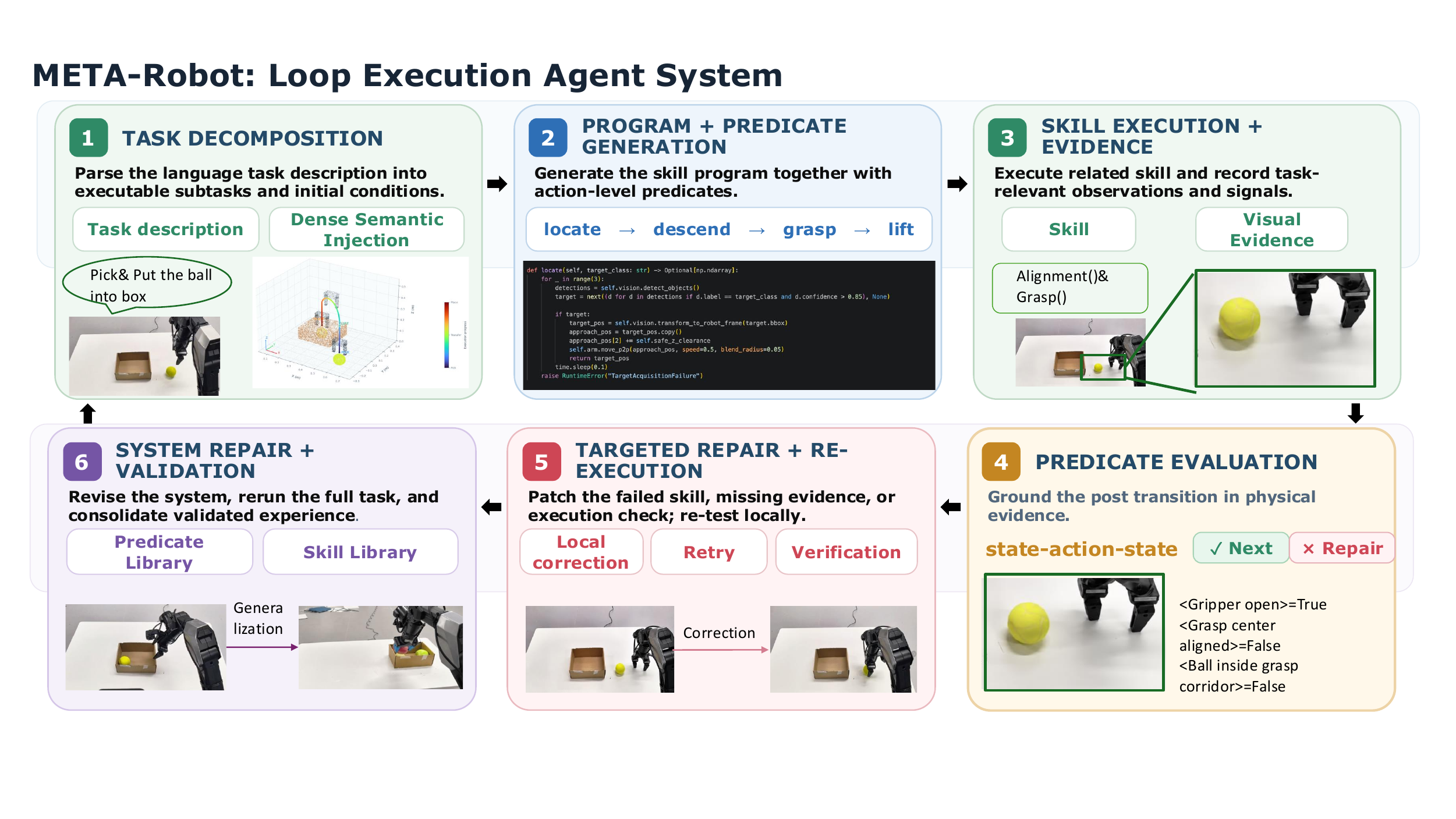}
    \caption{META-Robot Loop Execution Agent. Given a task description, the agent (1) decomposes it into executable subtasks and initial conditions; (2) jointly generates the task program and action-level predicates; (3) executes each skill while collecting task-relevant evidence; (4) evaluates predicate transitions to determine whether execution should continue or be repaired; (5) locally corrects and re-executes the failed skill, missing evidence routine, or verification check; and (6) performs system-level repair and full-task validation, consolidating validated predicates and skills for reuse. This procedure is repeated to form a closed execution--evaluation--repair loop.}
    \label{fig:META-Robot Loop Execution Agent}
\end{figure}

Inspired by software coding agents \cite{deepseek-harness2026,barbaste2026harness}, we treat evaluation of the execution system itself as a key component of scaling, just as shown in Fig.\ref{fig:META-Robot Loop Execution Agent}. Existing robotic agents often focus on task outcomes or intermediate execution feedback, with limited assessment of the mechanisms responsible for observation and evaluation. When failures arise beyond these mechanisms' coverage---the system's cognition and capability boundaries---repeated iteration alone may be insufficient.  To solve this problem, META-Robot's loop execution agent system uses action-level predicates to assess the outcome of each action execution. Given a task description, dense semantic injection \cite{smith2025steer} enriches its specification and decomposes it into a subtask trace. For each subtask, the agent generates action-level predicates from the task description and the LLM's own reasoning, establishing explicit criteria for execution assessment. These predicates are evaluated after each skill invocation and at subtask completion, providing fine-grained evidence of whether the intended action conditions have been satisfied.

To recover from an unsatisfied action-level condition, we initiate a bounded local repair loop whenever a required action-level predicate evaluates to False. To restore the expected state transition within a limited number of turns, the repair agent revises the code surrounding the failed action or applies iterative refinement procedures, like Picard iteration \cite{ince1956ordinary} and multi-directional approach search \cite{panic2026directional,ortega2000iterative}. To preserve diagnostic evidence when the failure cannot be localized or resolved locally, we record the latest predicate state and continue the remaining execution. To support task-level failure attribution, we consolidate all evaluations into a predicate transition trace. If the overall task fails, a global repair agent analyzes this trace together with execution logs and auxiliary evidence, formulates diagnostic hypotheses, and revises the task program or its supporting observation and verification mechanisms. To validate these system-level modifications, we re-execute the complete task, enabling improvements beyond isolated action-level repairs.

\subsection{Action-level predicate Library \& Skill Library}
\label{sec:skill_library}
META-Robot maintains two coupled libraries: a Skill Library that stores reusable execution knowledge distilled from validated repairs, and an action-level predicate library that stores verifiable action predicates. Skills specify how to execute or repair an interaction, while associated predicates specify the conditions to be satisfied after execution. Predicates and skills are associated: each skill can be associated with multiple predicates, and different skills may share the same predicate.

\paragraph{Formal representation.}
To couple reusable skills with action-level evaluation criteria, we represent the Skill Library as a forest
$\mathcal{L}_{S}=(\mathcal{V},\mathcal{E})$, where edges encode parent--child specialization. Each skill is defined as
\begin{equation}
s_v=
\left(
f_v,\mathcal{C}_v,g_v,r_v,
\mathcal{Q}_v,\pi_v,\mathcal{D}_v
\right),
\end{equation}
where $(f_v,\mathcal{C}_v,g_v,r_v)$ specify its failure signature, applicability, intent, and repair mechanism, while $(\mathcal{Q}_v,\pi_v,\mathcal{D}_v)$ record its feedback requirements, implementation, and validation evidence.

To organize evaluation criteria consistently, the Predicate Library contains shared action-level predicates
$\mathcal{P}=\{\phi_j\}_{j=1}^{M}$ and follows the same categorization as the Skill Library. Their correspondence is
\begin{equation}
\Gamma(v)\subseteq\mathcal{P},
\qquad
\mathcal{R}
=
\{(v,\phi_j)\mid\phi_j\in\Gamma(v)\},
\label{eq:skill_predicate_relation}
\end{equation}
allowing each skill to use multiple predicates and each predicate to be shared across skills.

To capture the effects of a skill invocation, the agent records
\begin{equation}
\mathbf{z}_{v,t}^{-}
=
[\phi_j(e_t^{-})]_{\phi_j\in\Gamma(v)},
\qquad
\mathbf{z}_{v,t}^{+}
=
[\phi_j(e_t^{+})]_{\phi_j\in\Gamma(v)},
\label{eq:skill_predicate_transition}
\end{equation}
where $e_t^{-}$ and $e_t^{+}$ denote evidence collected before and after execution. These transitions associate reusable skills with explicit changes in task conditions.
\paragraph{Parent--child organization and shared predicate indexing.}
To generalize validated repairs without imposing a predefined taxonomy, META-Robot organizes skills into parent--child relations. Parent nodes capture general repair principles, while child nodes specialize them to particular tasks, objects, sensing conditions, or embodiments. Let $\mathcal{A}(v)$ denote the contexts satisfying $\mathcal{C}_v$. A parent $p$ and child $c$ satisfy
\begin{equation}
\mathcal{A}(c)\subseteq\mathcal{A}(p),
\qquad
r_c=\operatorname{Specialize}(r_p;\delta_c),
\label{eq:skill_specialization}
\end{equation}
where $\delta_c$ introduces additional conditions or implementation details. This relation represents semantic specialization rather than guaranteed success over the entire parent domain.

To align evaluation criteria with skill specialization, the Predicate Library is indexed by the same hierarchy. For a skill category $b$,
\begin{equation}
\mathcal{P}_{b}=\bigcup_{v\in\operatorname{Subtree}(b)}\Gamma(v)
\label{eq:predicate_category}
\end{equation}
To avoid duplication, predicates with identical semantics and evidence requirements are shared across skills and branches. A specialized skill inherits applicable predicates and introduces new ones only when the existing predicate set cannot characterize the interaction. For example, torque-guided thin-towel grasping may specialize contact-aware grasping, reuse contact detected, and introduce cloth elevated and grasp verified. Contact and successful acquisition remain separate predicates because they describe distinct physical conditions.
\paragraph{Incremental consolidation.}
To convert validated repairs into reusable capabilities, the agent distills each repair $\rho$ into a candidate skill $\hat{s}$ and compares its failure signature, applicability conditions, execution intent, and corrective mechanism with related library entries. Equivalent repairs are merged, specialized repairs become child nodes, and distinct mechanisms create new branches. Changes to applicability or implementation semantics are revalidated before admission.

To preserve consistent evaluation, the agent jointly consolidates the predicates associated with the new skill. Existing predicates are reused when their semantics and evidence requirements match; otherwise, new predicates are proposed, validated, and added to $\mathcal{P}$. The corresponding mapping $\Gamma(v)$ is then updated. Reorganizing the skill hierarchy changes only predicate indexing, while shared predicate identities remain unchanged.

To transfer validated knowledge to a new context $x$, the agent retrieves the most specific applicable skill whose feedback requirements are available:
\begin{equation}
v^\star \in
\arg\max_{\substack{
v\in\mathcal{V}\\
x\in\mathcal{A}(v)\\
\mathcal{Q}_v\subseteq\mathcal{Q}(x)
}}
\operatorname{depth}(v).
\label{eq:skill_retrieval}
\end{equation}
The selected skill is retrieved together with $\Gamma(v^\star)$, providing both execution guidance and action-level evaluation criteria. If no specialized skill applies, a compatible parent guides the generation and validation of a new implementation. The whole algorithm flow pseudocode can be seen in appendix \ref{app:Predicae &Skill library}
\subsection{proprioceptive signal orchestration mechanism  }
To enable the agent to reason about the robot's own perceptual capabilities, META-Robot introduces a mechanism for the synergistic use of proprioceptive signals. Instead of relying on fixed observation interfaces, the system identifies available proprioceptive signals---such as joint states, motor currents, force/torque measurements, and contact-related feedback---by examining specifications tied to the robot's embodiment (e.g., simulator configurations, controller parameters, and robot description files like URDF). The system then associates these signals with their physical semantics and utilizes them as actionable sources of evidence during execution.

\begin{figure}[!htbp]
    \centering
    \includegraphics[width=1\linewidth]{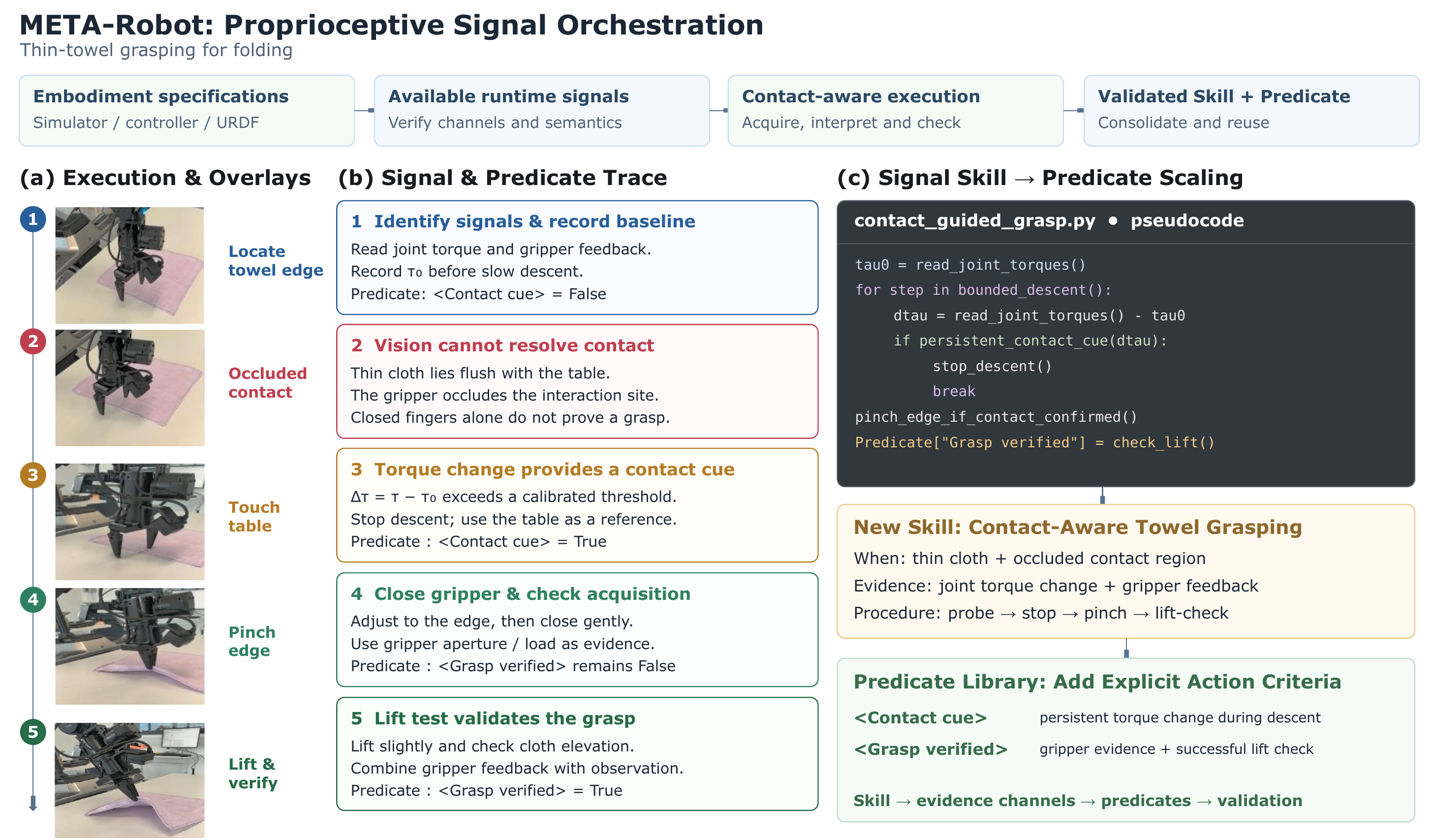}
    \caption{Proprioceptive Signal Orchestration Mechanism Example: in thin-towel grasping, torque change provides a contact cue, while gripper feedback and a lift check verify successful acquisition to complement insufficient visual evidence during execution and verification. Validated signal usage is then consolidated into a reusable skill together with corresponding action predicates, jointly expanding sensing, execution, and verification capabilities}
    \label{fig:PROPRIOCEPTIVE SIGNAL ORCHESTRATION MECHANISM Example}
\end{figure}

When a task indicates that visual information alone is insufficient to achieve the goal, the system incorporates these identified signals into the execution and verification processes. Validated proprioceptive applications are refined and categorized within the skill library (e.g., contact-aware grasping, slip detection, or force-sensitive manipulation), while the predicate library is expanded to include corresponding action-level criteria that define the state conditions required before and after an interaction. For instance, the assessment of a grasping action is no longer limited to a visual check of the gripper's closed state; it can also be determined based on contact conditions, motor load, or force stability. This is particularly crucial for failure cases that are visually ambiguous (such as "near-miss" grasps) or for tasks involving deformable and thin objects (like folding a towel, just as shown in Fig.\ref{fig:PROPRIOCEPTIVE SIGNAL ORCHESTRATION MECHANISM Example}), where image evidence alone often fails to confirm whether stable contact or effective manipulation has been achieved.

Through this process, META-Robot moves beyond treating proprioception merely as a static auxiliary input; instead, it continuously explores how embodiment-specific signals can support novel execution strategies and verification criteria, thereby improving the cognition and capabilities of the system. As these signal-based skills and predicate-based criteria accumulate, the system gains access to interaction state information that is difficult to infer visually, leading to significant improvements in meta-level generalization to more complex tasks.

\section{Experiments}
\subsection{Experimental Setup}
We evaluated META-Robot on LIBERO-Pro \cite{zhou2025liberopro} with $\pi_0$ \cite{black2024pi_0}, $\pi_{0.5}$, \cite{intelligence2025pi_}, OpenVLA \cite{kim2024openvla}, etc,  in simulation environment and on the Galaxea R1 Lite \cite{galaxeaR1Lite}  platform in real-world experiments. For simulation, we used GPT-6 Astra \cite{openai2026gpt6astra} and DeepSeek-V4-Flash \cite{xu2026deepseek} as coding agent. We tested several different code models to support, please see appendix \ref{app:Predicae &Skill library}. Given a natural-language task description, the agent generates executable Python programs and iteratively refines them using action-level predicate state-transition traces. The predicate and skill libraries provide reusable action evaluation criteria and implementations to support policy generation and repair.

For real-world experiments, we use GPT-5.6-sol \cite{openai2026gpt56sol} on R1 Lite. Our tasks cover basic pick-and-place, sequential multi-object manipulation, instruction following, precise plug insertion, thin-towel folding, and breakfast-item stacking. We additionally consider identifying an empty bottle among three visually identical opaque bottles containing different amounts of water, using joint-torque feedback to distinguish their loads. The system combines head-mounted camera observations with available proprioceptive signals, including joint torques and gripper feedback. These signals support interaction judgments that are difficult to establish visually, such as detecting contact with the tabletop when grasping a thin towel. Across both settings, META-Robot evaluates execution through predicate checks after skill calls and subtasks, attempts bounded local repairs, and uses accumulated predicate traces to guide system-level revision and full-task re-execution. Adaptation is performed through program and library updates without model parameter training.

\subsection{Main Evaluation Results}
Tab. ~\ref{tab:libero_pro_results} compares META-Robot with prior methods across the Object, Goal, and Spatial suites under position and task perturbations, evaluated over 50 seeds. META-Robot achieves its largest gains on Spatial, where the GPT-6 variant reaches 89.25\% on Pos and 95.8\% on Task, outperforming ASPIRE by 38.25 and 35.8 percentage points, respectively. On Goal--Task, DeepSeek-V4-Flash and GPT-6 achieve 64.6\% and 69.8\%, compared with ASPIRE's 45.0\%. Performance also remains strong on Object, with GPT-6 reaching 99.6\% on Pos and 99.0\% on Task. On Goal--Pos, META-Robot with GPT-6 achieves 85.8\%, outperforming ASPIRE's 81.0\% by 4.8 percentage points. Overall, both META-Robot variants perform competitively across the benchmark, while GPT-6 consistently achieves the stronger results, particularly on spatial perturbations and Goal--Task. Per-task success rates are reported in the appendix.

\begin{table}[!htbp]
\centering
\caption{LIBERO-Pro evaluates the success rate of each model across Object, Goal, and Spatial tasks}
\label{tab:libero_pro_results}

\small
\setlength{\tabcolsep}{2.4pt}
\renewcommand{\arraystretch}{1.25}

\resizebox{\textwidth}{!}{

\begin{tabular}{@{}llcccccccc@{}}
\toprule
\textbf{Category} & \textbf{\shortstack{Task\\Type}} & \textbf{OpenVLA} & \textbf{$\pi_0$} & \textbf{$\pi_{0.5}$} & \textbf{\shortstack{CaP-\\Agent0}} & \textbf{ASPIRE} & \textbf{\shortstack{Playful /\\RATS}} & \textbf{\shortstack{META-Robot\\(DeepSeek-\\V4-Flash)}} & \textbf{\shortstack{META-Robot\\(GPT-6 Astra)}} \\
\midrule

\multirow{2}{*}{\textbf{Object}}
  & Pos  & 0.00 & 0.00 & 17.00 & 22.00 & 98.00 & 61.00 & 98.00 & 99.60 \\
  & Task & 0.00 & 0.00 & 1.00  & 18.00 & 95.00 & 63.00 & 88.60 & 99.00 \\
\midrule
\multirow{2}{*}{\textbf{Goal}}
  & Pos  & 0.00 & 0.00 & 38.00 & 26.00 & 81.00 & 43.00 & 72.20 & 85.80 \\
  & Task & 0.00 & 0.00 & 0.00  & 17.00 & 45.00 & 36.00 & 64.60 & 69.80 \\
\midrule
\multirow{2}{*}{\textbf{Spatial}}
  & Pos  & 0.00 & 0.00 & 20.00 & 12.00 & 51.00 & 29.00 & 83.00 & 89.25 \\
  & Task & 0.00 & 0.00 & 1.00  & 14.00 & 60.00 & 31.00 & 88.40 & 95.80 \\
\bottomrule
\end{tabular}
}
\end{table}

\subsection{Real-Robot Evaluation}

\begin{figure}[!htbp]
    \centering
    \includegraphics[width=1\linewidth]{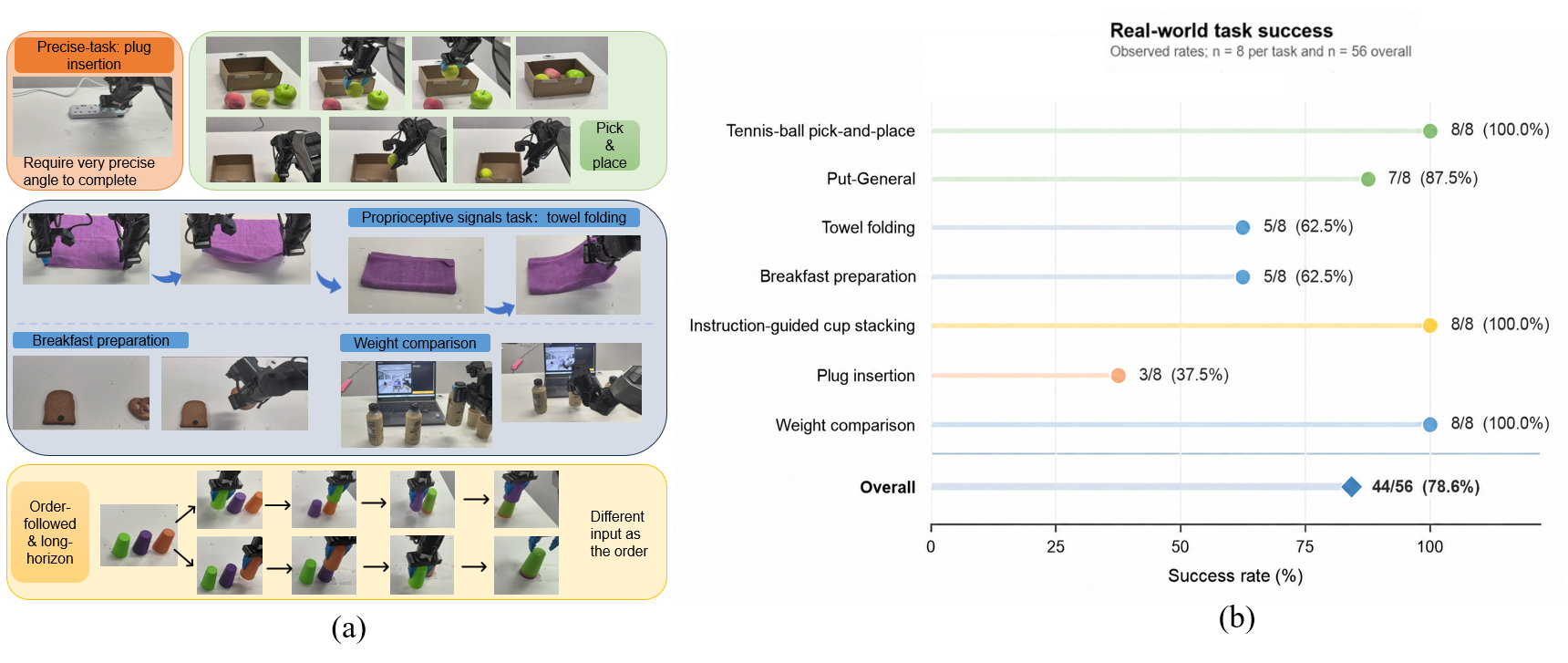}
    \caption{(a) illustrates the robot performing diverse real-world tasks, including precise insertion, pick-and-place, proprioceptive signals task and long-horizon instruction following. (b) shows the success rate of these tasks. }
    \label{fig:real-machine success}
\end{figure}

Fig.~\ref{fig:real-machine success} reports META-Robot's performance across seven real-world tasks (with eight trials per task) was as follows: the system succeeded in 44 out of 56 total trials (a success rate of 78.6\%). Specifically, the tennis ball pick-and-place, instruction-guided cup stacking, and weight comparison tasks all achieved a 100\% success rate, while the "Put-General" task had a success rate of 87.5\%. The success rates for towel folding and breakfast preparation dropped to 62.5\% (notably, due to the thin, soft, brittle, and slippery nature of towels and cookies, the success rate for these two tasks was 0\% without proprioceptive feedback, but improved significantly once such feedback was incorporated; see the appendix for a detailed analysis). The plug insertion task proved the most challenging, with a success rate of only 37.5\%. These results indicate that while the robot excels at tasks such as pick-and-place, instruction following, and weight comparison, manipulating deformable objects and achieving contact alignment remain its primary limitations. More details can be seen in \ref{app:The analysis ot each real machine task}.

\subsection{Zero-Shot Transfer to Unseen Tasks}

\begin{figure}[!htbp]
    \centering
    \includegraphics[width=0.68\linewidth]{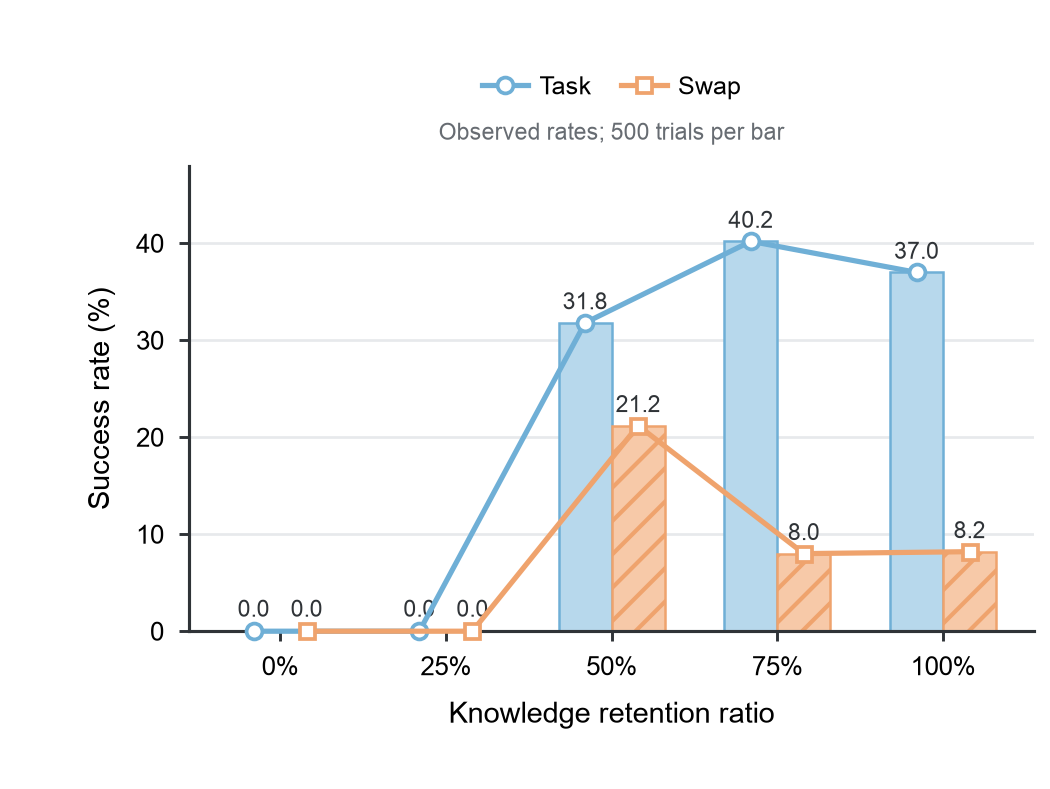}
    \caption{Zero-shot transfer to LIBERO-Pro Long under different library retention ratios. Success rates are averaged over ten tasks.}
    \label{fig:long_transfer_chart}
\end{figure}

We evaluate transfer to LIBERO-Pro Long using fixed predicate and skill libraries, without additional debugging or library updates, as shown in Fig.~\ref{fig:long_transfer_chart}. We randomly retain \{25\%, 50\%, 75\%, 100\%\} of the skills and their associated predicates. On Task perturbations, success is 0.0\%, 31.8\%, 40.2\%, and 37.0\% at the four retention ratios, respectively. The corresponding results on Swap perturbations are 0.0\%, 21.2\%, 8.0\%, and 8.2\%. These results show that accumulated predicate--skill knowledge supports partial zero-shot transfer, but its benefit varies substantially with the perturbation type and retained library composition. In particular, the non-monotonic Task trend and the weaker Swap results indicate that retaining more entries does not necessarily improve retrieval or execution. However, because each ratio is evaluated using a randomly retained subset, the experiment cannot determine whether the differences arise from redundant context, missing task-relevant entries, or sampling variation. Repeated subsampling across multiple library masks is therefore required before interpreting 75\% retention as an optimal trade-off. Per-task results in Figure. ~\ref{app:subtask analysis of zero-shot transfer} further show that gains are concentrated in a subset of tasks, suggesting that transfer depends on whether the retained skills and predicates cover each task's required conditions rather than on library size alone.

\FloatBarrier
\subsection{Ablation Studies}
\subsubsection{Action-level Predicate Ablation Test}
 As shown in Tab.~\ref{tab:libero_pro_fg_ablation}, we evaluate predicate contributions by removing them across all six LIBERO-Pro splits. On task perturbations, success falls from 99.0\% to 85.0\% for Object, 69.8\% to 56.0\% for Goal, and 95.8\% to 76.0\% for Spatial. On Pos perturbations, it falls from 99.6\% to 91.0\%, 85.8\% to 69.0\%, and 89.25\% to 72.0\%, respectively. These consistent drops show that action-level predicates support execution assessment, failure localization, and targeted repair. The Spatial-Pos comparison warrants caution because the full-system result covers eight of the available tasks.
\begin{table}[!htbp]
\centering
\small
\renewcommand{\arraystretch}{1.2}
\caption{Ablation Study of META-Robot across Different Tasks on LIBERO-Pro: We compare the task success rates (\%) of full META-Robot and its variant without action-level predicates across the Object, Goal, and Spatial suites under task and position perturbations. Removing predicates consistently reduces performance across all six settings, demonstrating their contribution to execution assessment and targeted repair.}
\label{tab:libero_pro_fg_ablation}
\vspace{0.3em}
\begin{tabular}{@{}llcc@{}}
\toprule
\textbf{Category} & \textbf{\shortstack{Task\\Type}} & \textbf{\shortstack{Full META-Robot\\(\%)}} & \textbf{\shortstack{Without Predicate\\(\%)}} \\
\midrule
\multirow{2}{*}{\textbf{Object}}
  & Task & 99.00 & 85.00 \\
  & Pos & 99.60 & 91.00 \\
\midrule
\multirow{2}{*}{\textbf{Goal}}
  & Task & 69.80 & 56.00 \\
  & Pos & 85.80 & 69.00 \\
\midrule
\multirow{2}{*}{\textbf{Spatial}}
  & Task & 95.80 & 76.00 \\
  & Pos & 89.25 & 72.00 \\
\bottomrule
\end{tabular}
\end{table}
\subsubsection{Proprioception Ablation test}
\begin{figure}[H]
\centering
    \centering
    \includegraphics[width=1\linewidth]{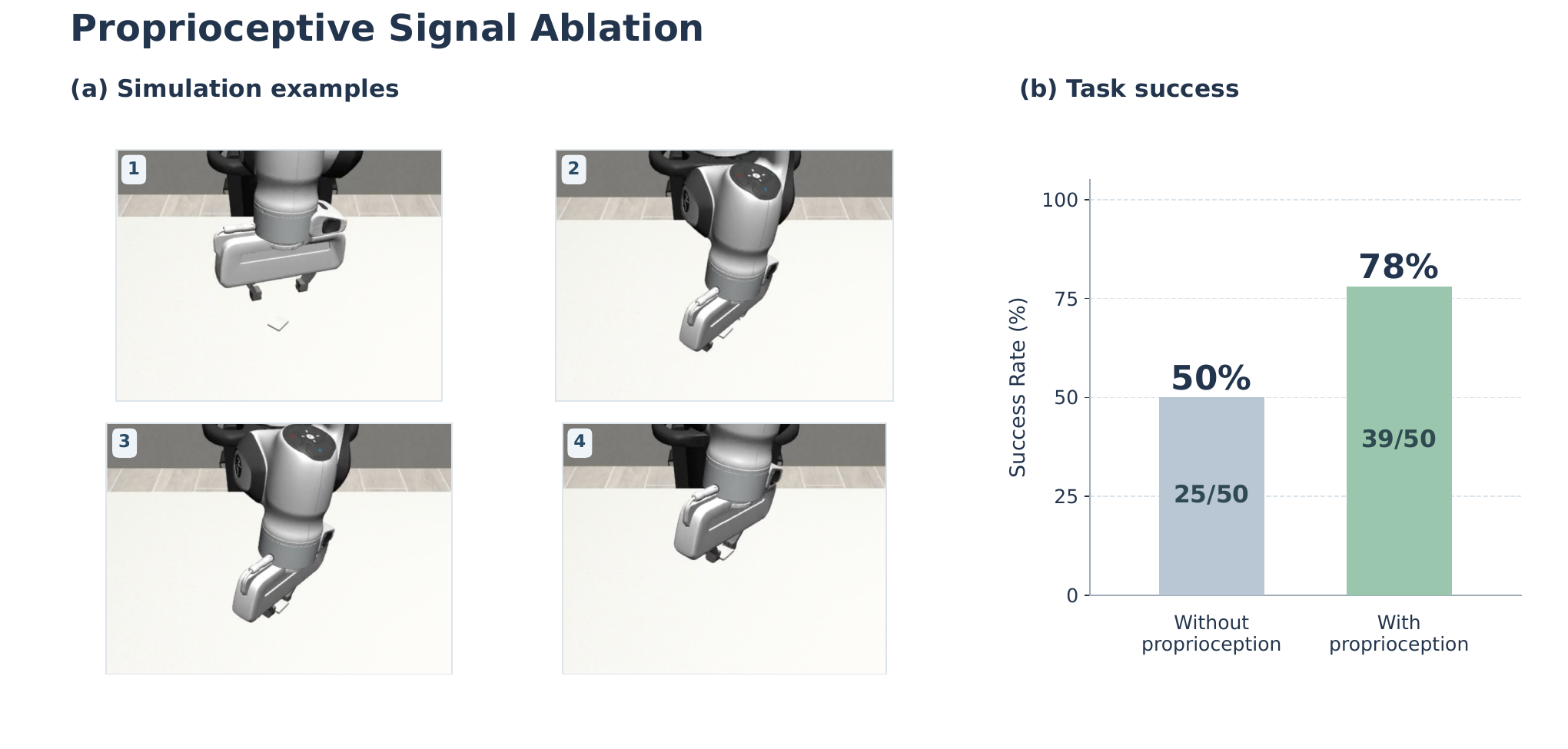}
    \caption{Effect of Proprioceptive Feedback on Manipulation Success: (a) shows the execution process during the simulation, while (b) displays the success rates before and after the ablation of the agent's signals.}
    \label{fig:META-Robot_Proprioceptive_Ablation}
\end{figure}

Libero-Pro lacks tasks that heavily rely on proprioceptive signals. Therefore, we specifically designed a new simulation task involving the grasping of a rigid, thin object for ablation testing; the success rate was 50\% without proprioceptive signals and 78\% with them, as shown in Fig.~\ref{fig:META-Robot_Proprioceptive_Ablation}.
The whole process can be divided into 4 steps: 1. The gripper positions itself and approaches the sheet; 2. it descends slowly to locate the point of contact; 3. it uses internal feedback to determine whether the gripper has made contact with the sheet or the tabletop and adjusts the grasping position; 4. it closes the gripper, lifts the sheet, and verifies the success of the grasp. Since the sheet lies flat against the tabletop, the contact point is difficult to determine based on visual data alone; therefore, this process requires assistance from internal system signals. In addition, we conducted a comparative analysis between the biscuit-grasping task performed during the real-world breakfast-making scenario and the task at hand; please refer to the appendix \ref{app:The analysis ot each real machine task} for further analysis and illustrations regarding this ablation study.

\FloatBarrier
\section{Conclusion}

This paper presented META-Robot, a self congnition and capbility scaling robotic agent that jointly improves execution, observation, and evaluation of task and system itself. Predicate state-transition traces support failure diagnosis and targeted repair, while coupled Predicate and Skill Libraries consolidate reusable criteria and procedures. Proprioceptive signal orchestration further incorporates physical evidence beyond vision. Experiments on LIBERO-Pro and the Galaxea R1 Lite demonstrate effective manipulation and transfer to held-out long-horizon tasks without additional debugging. The varying transfer performance also shows that more accumulated knowledge does not necessarily yield better generalization, emphasizing the importance of selecting task-relevant experience. More broadly, META-Robot supports joint cognitive and capability scaling by refining not only executable skills, but also the evidence channels and verification criteria used to assess them. These results motivate robotic self scaling through the coordinated refinement of execution, sensing, and evaluation. Future work will study relevance-aware knowledge selection, automatic signal calibration, and broader continual evaluation across tasks, embodiments, and contact-rich settings.
\label{headings}

\FloatBarrier

\begin{minipage}{\linewidth}
\textbf{AI use statement}

Generative AI was used to assist the research and writing workflow. We  used generative AI tools to assist with drafting and editing text, identifying relevant
literature, providing feedback on experimental design and
evaluation protocols, and examining the interpretation and
limitations of the reported results. The authors reviewed the
AI-assisted text and suggestions, checked cited sources and
empirical claims against the original publications and
experimental records, and determined which suggestions to
incorporate. The authors take full responsibility for the
methodology, implementation, reported results, references,
and conclusions of this work, including all content produced
with the assistance of generative AI.

\end{minipage}
\clearpage
\bibliographystyle{plainnat}
\bibliography{references}

\begin{thebibliography}{32}
\providecommand{\natexlab}[1]{#1}
\providecommand{\url}[1]{\texttt{#1}}
\expandafter\ifx\csname urlstyle\endcsname\relax
  \providecommand{\doi}[1]{doi: #1}\else
  \providecommand{\doi}{doi: \begingroup \urlstyle{rm}\Url}\fi

\bibitem[Athalye et~al.(2026)Athalye, Kumar, Silver, Liang, Wang,
  Lozano-P{\'e}rez, and Kaelbling]{athalye2026pixels}
Ashay Athalye, Nishanth Kumar, Tom Silver, Yichao Liang, Jiuguang Wang,
  Tom{\'a}s Lozano-P{\'e}rez, and Leslie~Pack Kaelbling.
\newblock From pixels to predicates: Learning symbolic world models via
  pretrained vlms.
\newblock \emph{IEEE Robotics and Automation Letters}, 2026.

\bibitem[Barbaste et~al.(2026)Barbaste, Darrigol, Vu, and
  Wiltberger]{barbaste2026harness}
Paul Barbaste, Tristan Darrigol, Germain Vu, and Tom Wiltberger.
\newblock Harness engineering: Anatomy, architecture, and evolution of coding
  agents--a source-code study of eleven systems.
\newblock \emph{arXiv preprint arXiv:2609.00006}, 2026.

\bibitem[Black et~al.(2024)Black, Brown, Driess, Esmail, Equi, Finn, Fusai,
  Groom, Hausman, Ichter, et~al.]{black2024pi_0}
Kevin Black, Noah Brown, Danny Driess, Adnan Esmail, Michael Equi, Chelsea
  Finn, Niccolo Fusai, Lachy Groom, Karol Hausman, Brian Ichter, et~al.
\newblock {$\pi_0$: A Vision-Language-Action Flow Model for General Robot
  Control}.
\newblock \emph{arXiv preprint arXiv:2410.24164}, 2024.

\bibitem[DeepSeek-AI(2026)]{deepseek-harness2026}
DeepSeek-AI.
\newblock Deepseek harness: Everything is a plugin.
\newblock \url{https://github.com/deepseek-ai/deepseek-harness}, 2026.

\bibitem[Elmaaroufi et~al.(2026)Elmaaroufi, Svegliato, Kalade, Schelle, Seshia,
  and Zaharia]{elmaaroufi2026rho}
Karim Elmaaroufi, Justin Svegliato, Sarunas Kalade, Graham Schelle, Sanjit~A
  Seshia, and Matei Zaharia.
\newblock Rho: Your coding agent is secretly a roboticist.
\newblock \emph{arXiv preprint arXiv:2606.16458}, 2026.

\bibitem[Fu et~al.(2026)Fu, Yu, El-Refai, Kou, Xue, Huang, Xiao, Wang, Niu, Li,
  et~al.]{fu2026cap}
Letian Fu, Justin Yu, Karim El-Refai, Ethan Kou, Haoru Xue, Huang Huang, Wenli
  Xiao, Guanzhi Wang, Dantong Niu, Fei-Fei Li, et~al.
\newblock Cap-x: A framework for benchmarking and improving coding agents for
  robot manipulation.
\newblock \emph{arXiv preprint arXiv:2603.22435}, 2026.

\bibitem[{Galaxea Dynamics}(n.d.)]{galaxeaR1Lite}
{Galaxea Dynamics}.
\newblock {R1 Lite} hardware introduction.
\newblock Galaxea Dynamics Documentation, n.d.
\newblock URL
  \url{https://deepwiki.com/userguide-galaxea/Product_User_Guide/2.3.1-r1-lite-hardware-introduction}.
\newblock Accessed September 26, 2026.

\bibitem[Ince(1956)]{ince1956ordinary}
Edward~L Ince.
\newblock \emph{Ordinary differential equations}.
\newblock Courier Corporation, 1956.

\bibitem[Intelligence et~al.(2025)Intelligence, Black, Brown, Darpinian,
  Dhabalia, Driess, Esmail, Equi, Finn, Fusai, et~al.]{intelligence2025pi_}
Physical Intelligence, Kevin Black, Noah Brown, James Darpinian, Karan
  Dhabalia, Danny Driess, Adnan Esmail, Michael Equi, Chelsea Finn, Niccolo
  Fusai, et~al.
\newblock {$\pi_{0.5}$: A Vision-Language-Action Model with Open-World
  Generalization}.
\newblock \emph{arXiv preprint arXiv:2504.16054}, 2025.

\bibitem[Kagaya et~al.(2025)Kagaya, Lakshmi, Lou, Yuan, Karlekar, Pranata,
  Murakami, Kinose, and You]{kagaya2025memorytransfer}
Tomoyuki Kagaya, Subramanian Lakshmi, Yuxuan Lou, Thong~Jing Yuan, Jayashree
  Karlekar, Sugiri Pranata, Natsuki Murakami, Akira Kinose, and Yang You.
\newblock Memory transfer planning: {LLM}-driven context-aware code adaptation
  for robot manipulation.
\newblock \emph{arXiv preprint arXiv:2509.24160}, 2025.

\bibitem[Kim et~al.(2024)Kim, Pertsch, Karamcheti, Xiao, Balakrishna, Nair,
  Rafailov, Foster, Lam, Sanketi, et~al.]{kim2024openvla}
Moo~Jin Kim, Karl Pertsch, Siddharth Karamcheti, Ted Xiao, Ashwin Balakrishna,
  Suraj Nair, Rafael Rafailov, Ethan Foster, Grace Lam, Pannag Sanketi, et~al.
\newblock Openvla: An open-source vision-language-action model, 2024.
\newblock \emph{URL https://arxiv. org/abs/2406.09246}, 1\penalty0
  (2):\penalty0 4, 2024.

\bibitem[Kumar(2026)]{kumar2026act}
Vaishak Kumar.
\newblock Act-observe-rewrite: Multimodal coding agents as in-context policy
  learners for robot manipulation.
\newblock \emph{arXiv preprint arXiv:2603.04466}, 2026.

\bibitem[Liang et~al.(2023)Liang, Huang, Xia, Xu, Hausman, Ichter, Florence,
  and Zeng]{liang2023code}
Jacky Liang, Wenlong Huang, Fei Xia, Peng Xu, Karol Hausman, Brian Ichter, Pete
  Florence, and Andy Zeng.
\newblock Code as policies: Language model programs for embodied control.
\newblock In \emph{2023 IEEE International conference on robotics and
  automation (ICRA)}, pages 9493--9500. IEEE, 2023.

\bibitem[Liang et~al.(2025)Liang, Kumar, Tang, Weller, Tenenbaum, Silver,
  Henriques, and Ellis]{liang2025visualpredicator}
Yichao Liang, Nishanth Kumar, Hao Tang, Adrian Weller, Joshua~B Tenenbaum, Tom
  Silver, Jo{\~a}o~F Henriques, and Kevin Ellis.
\newblock Visualpredicator: Learning abstract world models with neuro-symbolic
  predicates for robot planning.
\newblock In \emph{International Conference on Learning Representations},
  volume 2025, pages 60416--60444, 2025.

\bibitem[Liu et~al.(2026)Liu, Li, Yao, Shi, Zhou, Huang, Huang, and
  Mao]{liu2026guava}
Haowen Liu, Xirui Li, Shaoxiong Yao, Peng Shi, Tianyi Zhou, Jia-Bin Huang,
  Furong Huang, and Jiayuan Mao.
\newblock Guava: An effective and universal harness for embodied manipulation.
\newblock \emph{arXiv preprint arXiv:2606.18363}, 2026.

\bibitem[Lu et~al.(2026)Lu, Wu, Kou, Fu, Xiao, Mandlekar, Xu, Shi, Goldberg,
  Chen, et~al.]{lu2026aspire}
Runyu Lu, Yubo Wu, Ethan Kou, Letian Fu, Wenli Xiao, Ajay Mandlekar, Yinzhen
  Xu, Guanya Shi, Ken Goldberg, Ang Chen, et~al.
\newblock Aspire: Agentic/skills discovery for robotics.
\newblock \emph{arXiv preprint arXiv:2607.00272}, 2026.

\bibitem[Meng et~al.(2025{\natexlab{a}})Meng, Sun, Fest, Li, Bing, and
  Knoll]{meng2025growing}
Yuan Meng, Zhenguo Sun, Max Fest, Xukun Li, Zhenshan Bing, and Alois Knoll.
\newblock Growing with your embodied agent: A human-in-the-loop lifelong code
  generation framework for long-horizon manipulation skills.
\newblock \emph{arXiv preprint arXiv:2509.18597}, 2025{\natexlab{a}}.

\bibitem[Meng et~al.(2025{\natexlab{b}})Meng, Yao, Ye, Zhou, Zhang, Sun, Li,
  Bing, and Knoll]{meng2025embodied}
Yuan Meng, Xiangtong Yao, Haihui Ye, Yirui Zhou, Shengqiang Zhang, Zhenguo Sun,
  Xukun Li, Zhenshan Bing, and Alois Knoll.
\newblock Embodied long horizon manipulation with closed-loop code generation
  and incremental few-shot adaptation.
\newblock \emph{arXiv preprint arXiv:2503.21969}, 2025{\natexlab{b}}.

\bibitem[{OpenAI}(2026{\natexlab{a}})]{openai2026gpt56sol}
{OpenAI}.
\newblock {GPT-5.6 Sol} model.
\newblock OpenAI API documentation, 2026{\natexlab{a}}.
\newblock URL \url{https://developers.openai.com/api/docs/models/gpt-5.6-sol}.
\newblock Accessed September 26, 2026.

\bibitem[{OpenAI}(2026{\natexlab{b}})]{openai2026gpt6astra}
{OpenAI}.
\newblock {GPT-6 Astra} model.
\newblock OpenAI API documentation, 2026{\natexlab{b}}.
\newblock URL \url{https://developers.openai.com/api/docs/models/gpt-6-astra}.
\newblock Accessed September 26, 2026.

\bibitem[Ortega and Rheinboldt(2000)]{ortega2000iterative}
James~M Ortega and Werner~C Rheinboldt.
\newblock \emph{Iterative solution of nonlinear equations in several
  variables}.
\newblock SIAM, 2000.

\bibitem[Panic(2026)]{panic2026directional}
Stefan~R Panic.
\newblock Directional gradient and curvature approximation via legendre
  quadrature in unconstrained optimization.
\newblock \emph{Applied Numerical Mathematics}, 220:\penalty0 294--309, 2026.

\bibitem[Santos et~al.(2026)Santos, Khadraoui, Farhat, Yous, Teffahi, and
  Hacid]{santos2026alrm}
Vitor Gaboardi~dos Santos, Ibrahim Khadraoui, Ibrahim Farhat, Hamza Yous, Samy
  Teffahi, and Hakim Hacid.
\newblock Alrm: Agentic llm for robotic manipulation.
\newblock \emph{arXiv preprint arXiv:2601.19510}, 2026.

\bibitem[Shi et~al.(2025)Shi, Ichter, Equi, Ke, Pertsch, Vuong, Tanner,
  Walling, Wang, Fusai, et~al.]{shi2025hi}
Lucy~Xiaoyang Shi, Brian Ichter, Michael Equi, Liyiming Ke, Karl Pertsch, Quan
  Vuong, James Tanner, Anna Walling, Haohuan Wang, Niccolo Fusai, et~al.
\newblock Hi robot: Open-ended instruction following with hierarchical
  vision-language-action models.
\newblock \emph{arXiv preprint arXiv:2502.19417}, 2025.

\bibitem[Smith et~al.(2025)Smith, Irpan, Arenas, Kirmani, Kalashnikov, Shah,
  and Xiao]{smith2025steer}
Laura Smith, Alex Irpan, Montserrat~Gonzalez Arenas, Sean Kirmani, Dmitry
  Kalashnikov, Dhruv Shah, and Ted Xiao.
\newblock Steer: Flexible robotic manipulation via dense language grounding.
\newblock In \emph{2025 IEEE International Conference on Robotics and
  Automation (ICRA)}, pages 16517--16524. IEEE, 2025.

\bibitem[Wang et~al.(2026{\natexlab{a}})Wang, Zhao, He, Peng, and
  Zhang]{wang2026eeagent}
Jianzong Wang, Botao Zhao, Yayun He, Junqing Peng, and Xulong Zhang.
\newblock Evolvable embodied agent for robotic manipulation via long short-term
  reflection and optimization.
\newblock \emph{arXiv preprint arXiv:2604.13533}, 2026{\natexlab{a}}.

\bibitem[Wang et~al.(2026{\natexlab{b}})Wang, Zhang, Lin, Yuan, Lin, Mao, Fan,
  Gao, Dai, Li, et~al.]{wang2026embodiedskills}
Wei Wang, Wenqiao Zhang, Yutong Lin, Yuqian Yuan, Tianwei Lin, Jinhao Mao,
  Zhenxuan Fan, Mingjian Gao, Yang Dai, Wentong Li, et~al.
\newblock Embodiedskills: A unified framework for orchestrating, training, and
  deploying vla agents.
\newblock \emph{arXiv preprint arXiv:2609.01281}, 2026{\natexlab{b}}.

\bibitem[Xiao et~al.(2026)Xiao, Xie, Zhang, Lin, Fu, Xue, Lu, Yang, Dai, Wang,
  et~al.]{xiao2026enpire}
Wenli Xiao, Jia Xie, Tonghe Zhang, Haotian Lin, Letian Fu, Haoru Xue, Jalen Lu,
  Yi~Yang, Cunxi Dai, Zi~Wang, et~al.
\newblock Enpire: Agentic robot policy self-improvement in the real world.
\newblock \emph{arXiv preprint arXiv:2606.19980}, 2026.

\bibitem[Xu et~al.(2026)Xu, Lin, Xue, Wang, Xu, Wu, Zhang, Lin, Dong, Ling,
  et~al.]{xu2026deepseek}
Anyi Xu, Bangcai Lin, Bing Xue, Bingxuan Wang, Bingzheng Xu, Bochao Wu, Bowei
  Zhang, Chaofan Lin, Chen Dong, Chenchen Ling, et~al.
\newblock Deepseek-v4: Towards highly efficient million-token context
  intelligence.
\newblock \emph{arXiv preprint arXiv:2606.19348}, 2026.

\bibitem[Yuan et~al.(2025)Yuan, Bai, Fu, Zhou, Feng, Xu, Zhan, Karlsson, and
  Lu]{yuan2025being}
Haoqi Yuan, Yu~Bai, Yuhui Fu, Bohan Zhou, Yicheng Feng, Xinrun Xu, Yi~Zhan,
  B{\"o}rje~F Karlsson, and Zongqing Lu.
\newblock Being-0: A humanoid robotic agent with vision-language models and
  modular skills.
\newblock \emph{arXiv preprint arXiv:2503.12533}, 2025.

\bibitem[Zhang et~al.(2026)Zhang, Ge, Yoo, Fu, Yang, Liu, Saravanan, Yin, Yu,
  Niu, et~al.]{zhang2026playful}
Junyi Zhang, Jiaxin Ge, Hanjun Yoo, Letian Fu, Zihan Yang, Yaowei Liu, Raj
  Saravanan, Shaofeng Yin, Justin Yu, Dantong Niu, et~al.
\newblock Playful agentic robot learning.
\newblock \emph{arXiv preprint arXiv:2606.19419}, 2026.

\bibitem[Zhou et~al.(2025)Zhou, Xu, Tie, Chen, Zhang, Chu, Zhou, and
  Sun]{zhou2025liberopro}
Xueyang Zhou, Yangming Xu, Guiyao Tie, Yongchao Chen, Guowen Zhang, Duanfeng
  Chu, Pan Zhou, and Lichao Sun.
\newblock {LIBERO-PRO}: Towards robust and fair evaluation of
  vision-language-action models beyond memorization.
\newblock \emph{arXiv preprint arXiv:2510.03827}, 2025.

\end{thebibliography}

\clearpage
\appendix
\section{Appendix}
\subsection{The analysis of each real-robot task}
\label{app:The analysis ot each real machine task}
\begin{figure}[H]
\centering
    \includegraphics[width=.85\linewidth]{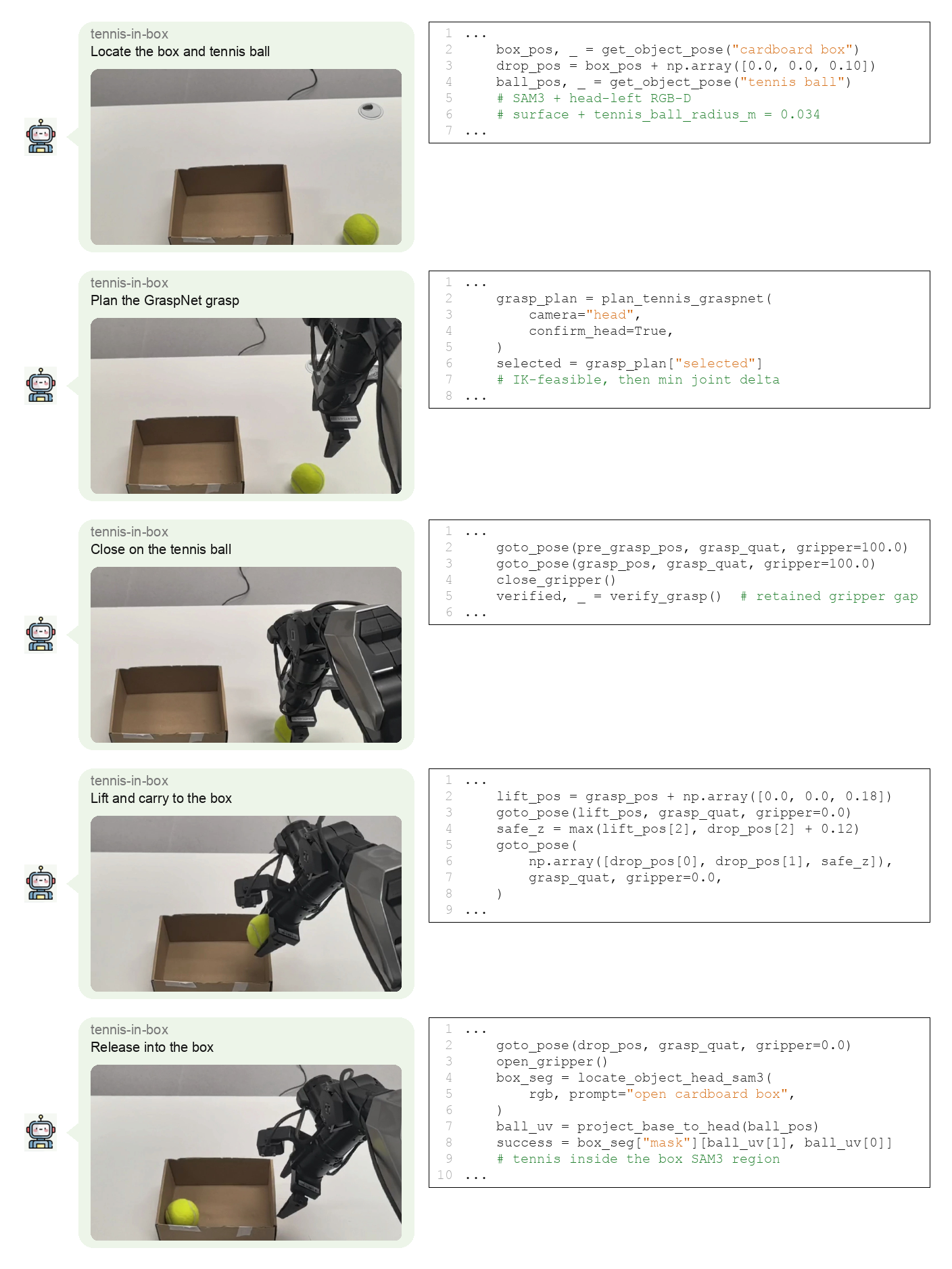}
    \caption{We validate the pick-and-place ability of META-Robot on a real-world Galaxea R1 Lite. A yellow tennis ball and an open cardboard box sit on the table; only the right arm moves. The operator names neither a grasp nor a drop point. We find META-Robot can easily complete this task. We find that META-Robot can pick and place from two live object poses rather than from a saved trajectory or large training data.}
    \label{fig:tennis pick-and-place.}
\end{figure}

META-Robot decomposes the task into locate, plan, pinch, carry, and verify. Head RGB-D and SAM3 recover the box and the ball; the ball center is the stereo surface plus the 3.4 cm radius. The drop pose is 10 cm above the box center. The agent then selects a grabbing position which first passes the IK harness with the smallest joint motion. The arm goes to pre-grasp, then grasp, and closes. The agent accepts the grasp only if the gripper gap is neither empty-closed nor fully open. The loaded arm lifts 18 cm, crosses at a safe height, and opens at the drop pose. After that, it checks that the ball lies within the box.\par

\clearpage
\begin{figure}[H]
\centering
    \includegraphics[width=.97\linewidth]{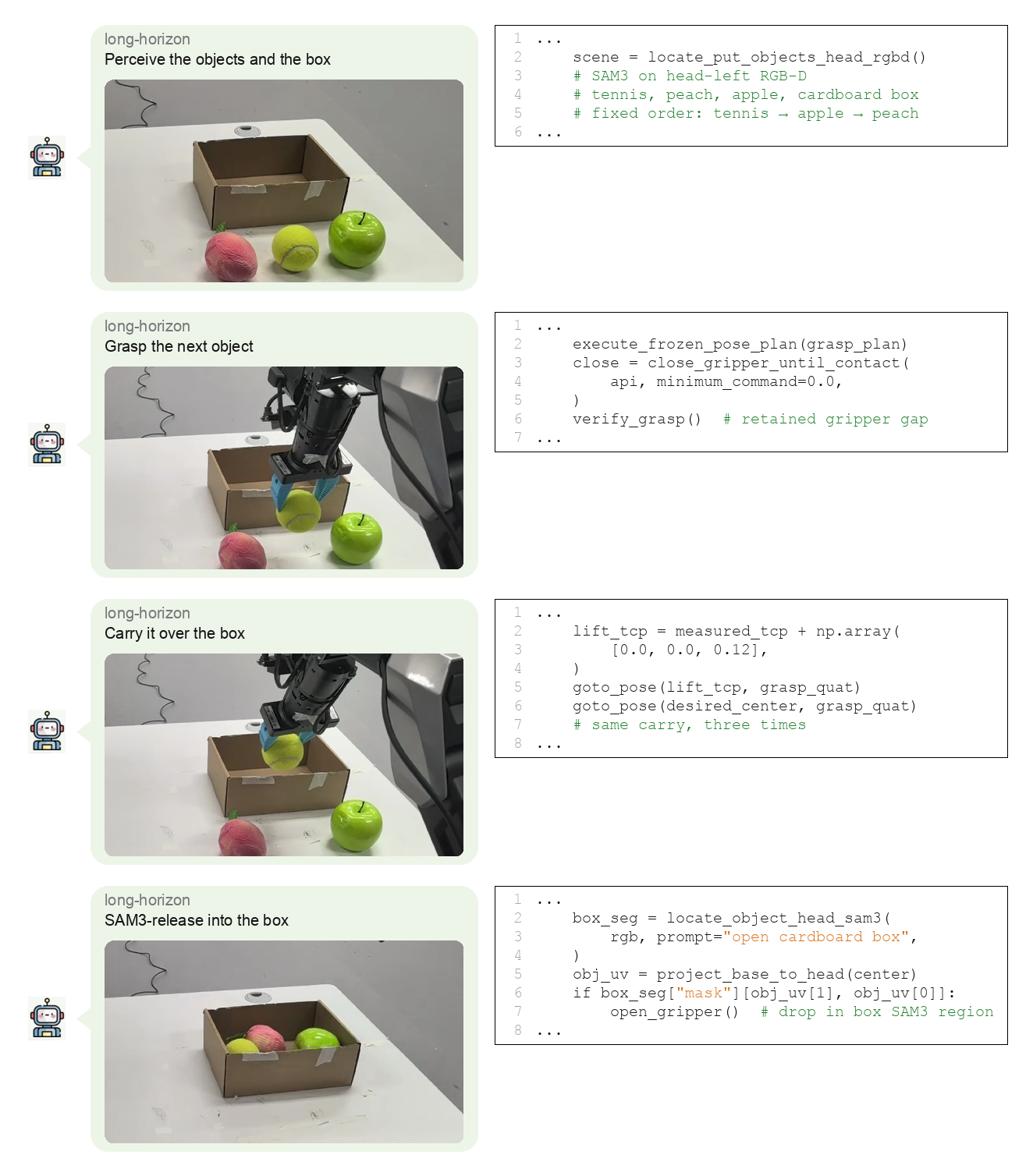}
    \caption{We validate META-Robot's long-horizon manipulation capability on a real-world Galaxea R1 Lite. A peach, a tennis ball, an apple, and an open cardboard box sit on the table; only the right arm moves. This task shows the robot's ability to perform long-horizon pick-and-place tasks.}
    \label{fig:put-general}
\end{figure}
This demo uses the tennis ball, followed by the apple and then the peach. No object is assigned a fixed spatial slot; only the order is specified. The robot is instructed to pick up the items and place them into the box in this order.\par
META-Robot decomposes the task into read-order, perceive, grasp, carry, and SAM3-release, repeated three times. One head-left RGB-D frame and SAM3 recover the three objects and the box. Each phase pinches the next object, verifies a retained gripper gap, and carries it over the box. After the third drop, all three objects sit in that same region.\par

\clearpage
\begin{figure}[H]
\centering
    \includegraphics[width=.96\linewidth]{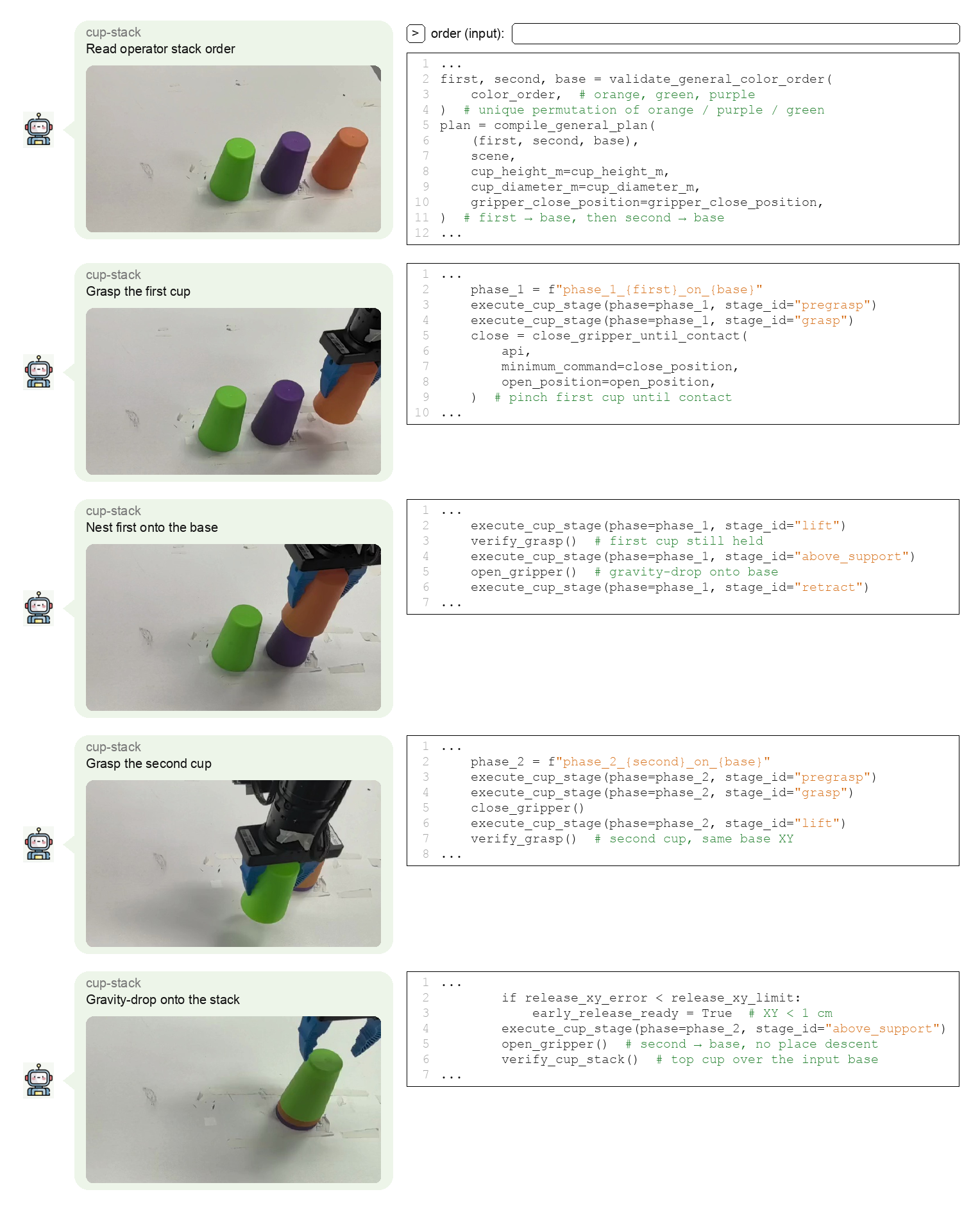}
    \caption{We validate the order-following ability of Meta-Robot on a real-world Galaxea R1 Lite using three inverted plastic cups of the same size---orange, purple, and green. The focus of this task is the robot's ability to follow the specified order, rather than the individual grasping and transfer motions. This real-world task evaluates generalization across different specified color orders.}
    \label{fig:Instruction-guided cup stacking}
\end{figure}
The cups already sit rim-down, so they nest rather than pile. We specify a unique color order (first, second, base): the first cup is placed onto the base, then the second cup is placed onto the same base. The three cups are initially arranged in a line in a random color order. Each time we input an order of the three colors.\par
META-Robot decomposes the task into read-order, perceive, grasp, transfer, and gravity-release. Head RGB-D and SAM3, gated by HSV, recover each inverted cup; the upward base is measured from stereo depth, and the downward rim is inferred from the known cup height. After compilation, phase 1 is first-on-base and phase 2 is second-on-base; both releases target the original base X-Y, because nested inverted cups keep the same rim height.\par

\clearpage
\begin{figure}[H]
\centering
    \includegraphics[width=.90\linewidth]{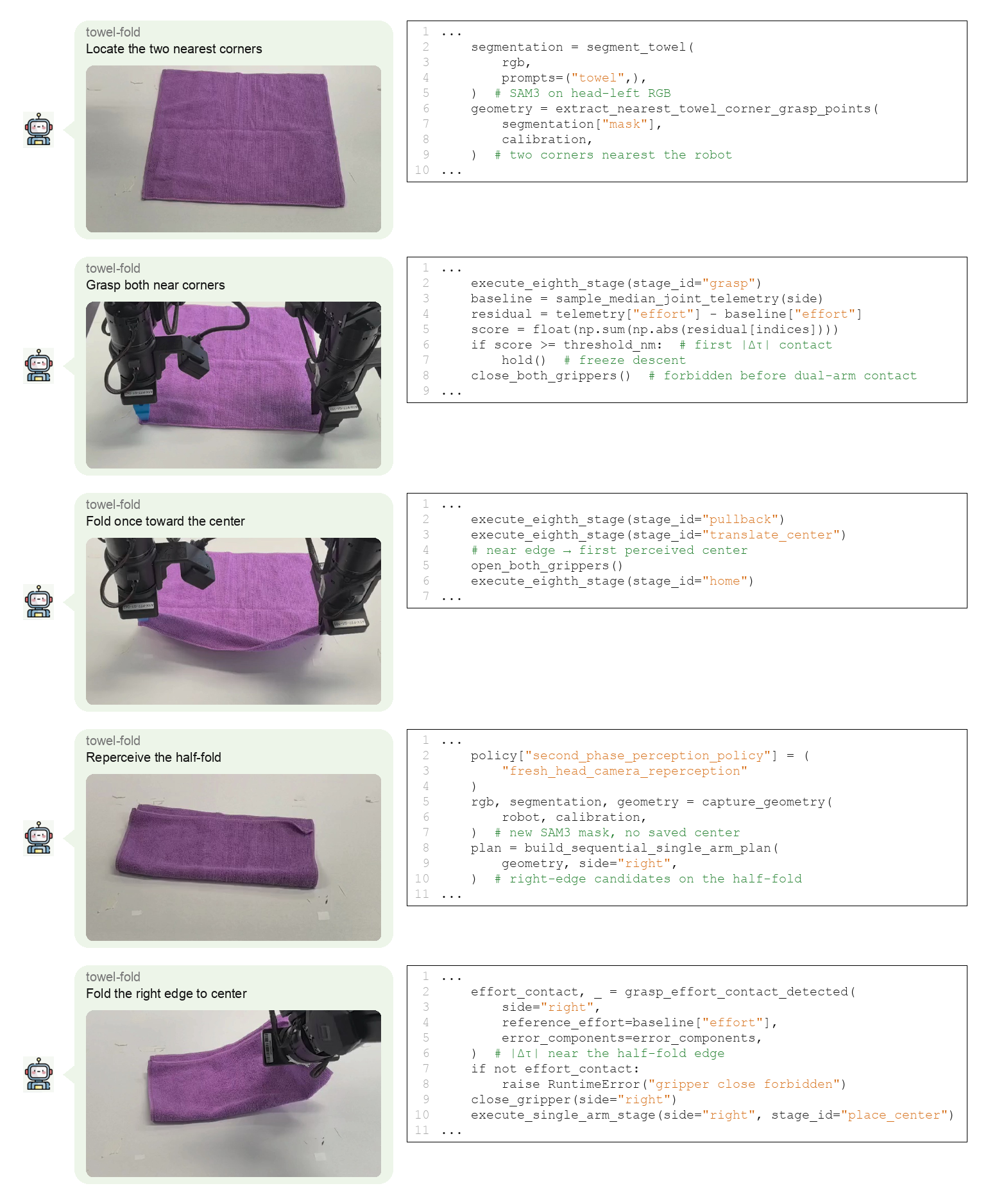}
    \caption{We validate the ability of META-Robot to fold a soft object on a real-world Galaxea R1 Lite using one flat purple cloth on the table. This task requires the use of proprioceptive signals as the towel is very thin and is quite difficult to grasp using visual feedback alone. META-Robot gives the plan: a bimanual first fold that turns the square into a half-width rectangle, then a right-arm second fold that turns that rectangle into a quarter. Both arms move in the first half; only the right arm moves in the second. The first episode makes the half-fold; the second episode makes the quarter. We validate its ability to handle complex tasks.}
    \label{fig:Towel folding}
\end{figure}

META-Robot decomposes the task into perceive-corners, bimanual-grasp, first-fold, reperceive, and second-fold. Head RGB and SAM3 recover the largest towel mask; a fitted rectangle's four corners are projected onto the real external contour, ranked by XY distance to the robot base, and the nearest pair becomes the left and right grasps. After the first fold changes the cloth's shape, the original corner pair is discarded. A fresh head frame re-segments the half-fold; META-Robot now treats it as a new rectangle and samples grasp candidates only on the current right edge, not the original four corners.\par
The core of this task is to verify its ability to deal with soft, thin objects, which requires the power of perception. One other important thing is to accurately grasp the edge of the towel and ensure there are no small areas of folding meanwhile. META-Robot uses the signal of joint torques to assist in determining the appropriate contact between the gripper and the table surface.\par

\clearpage
\begin{figure}[H]
\centering
    \centering
    \includegraphics[width=1\linewidth]{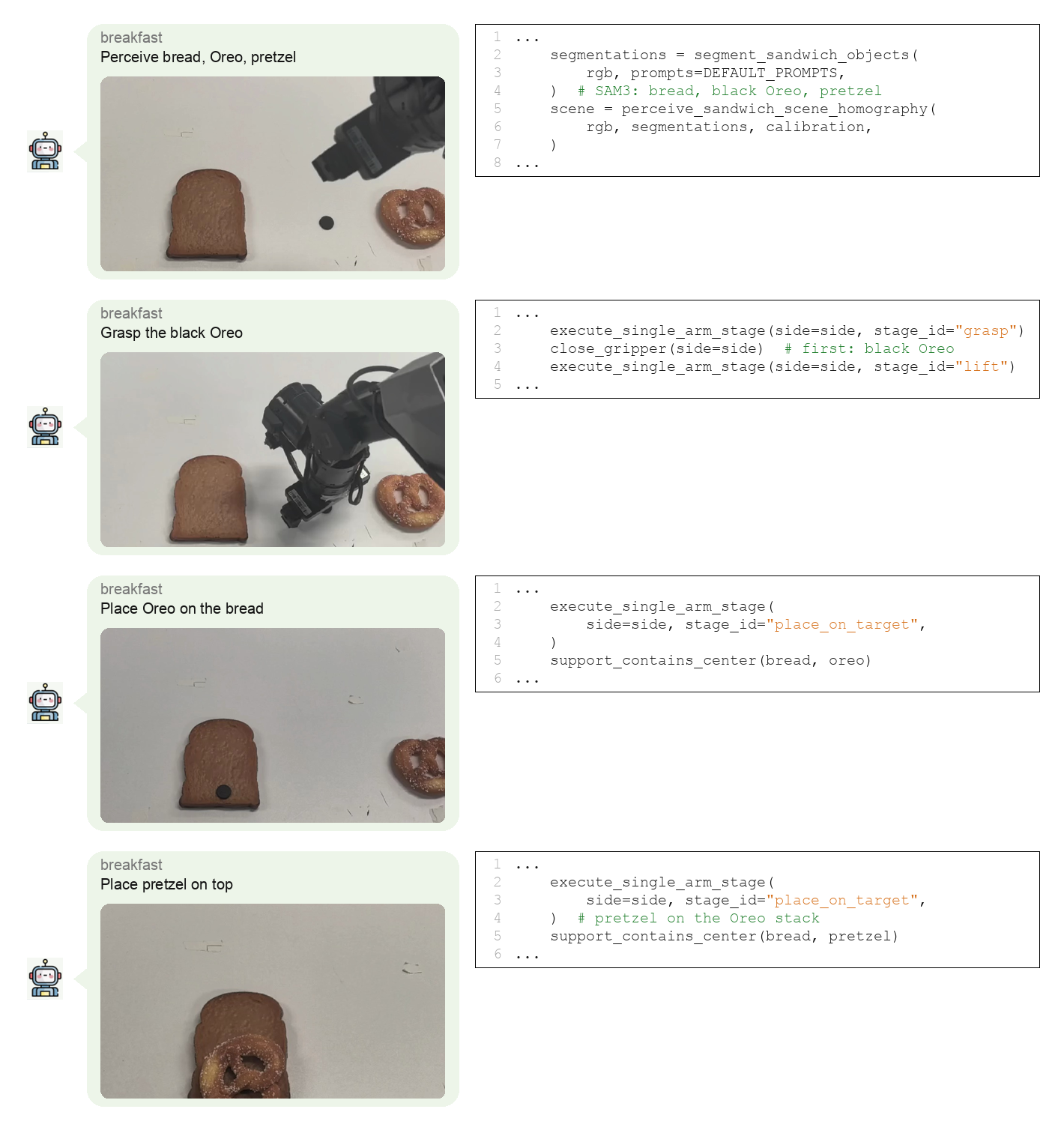}
    \caption{We validate the breakfast task of META-Robot on a real-world Galaxea R1 Lite. A brown toast slice, a small black Oreo, and a pretzel sit on the table; only one arm moves. We use a fixed two-layer order: Oreo onto the bread, then pretzel on top of that stack. This task is also an evaluation of the system's use of proprioceptive feedback.}
    \label{fig:Breakfast preparation}
\end{figure}

META-Robot decomposes the task into perceive, grasp, place-on-support, and verify, once per layer. One RGB frame and SAM3 recover the bread, the Oreo, and the pretzel to detect their positions. The first phase pinches the black Oreo and places it onto the toast. After a fresh re-perception, the second phase pinches the pretzel and places it on top of the Oreo.\par

In fact, the sub-task of grasping a cookie for breakfast shares similarities with the thin-wafer grasping task we simulated; the challenge in both cases lies in determining how to grasp objects whose grasp points are difficult to pinpoint using vision alone. While both tasks rely on detecting contact between the gripper and the table surface, grasping a cookie introduces the additional requirement of controlling the gripper's closing force.\par

\clearpage
\begin{figure}[H]
\centering
    \centering
    \includegraphics[width=.95\linewidth]{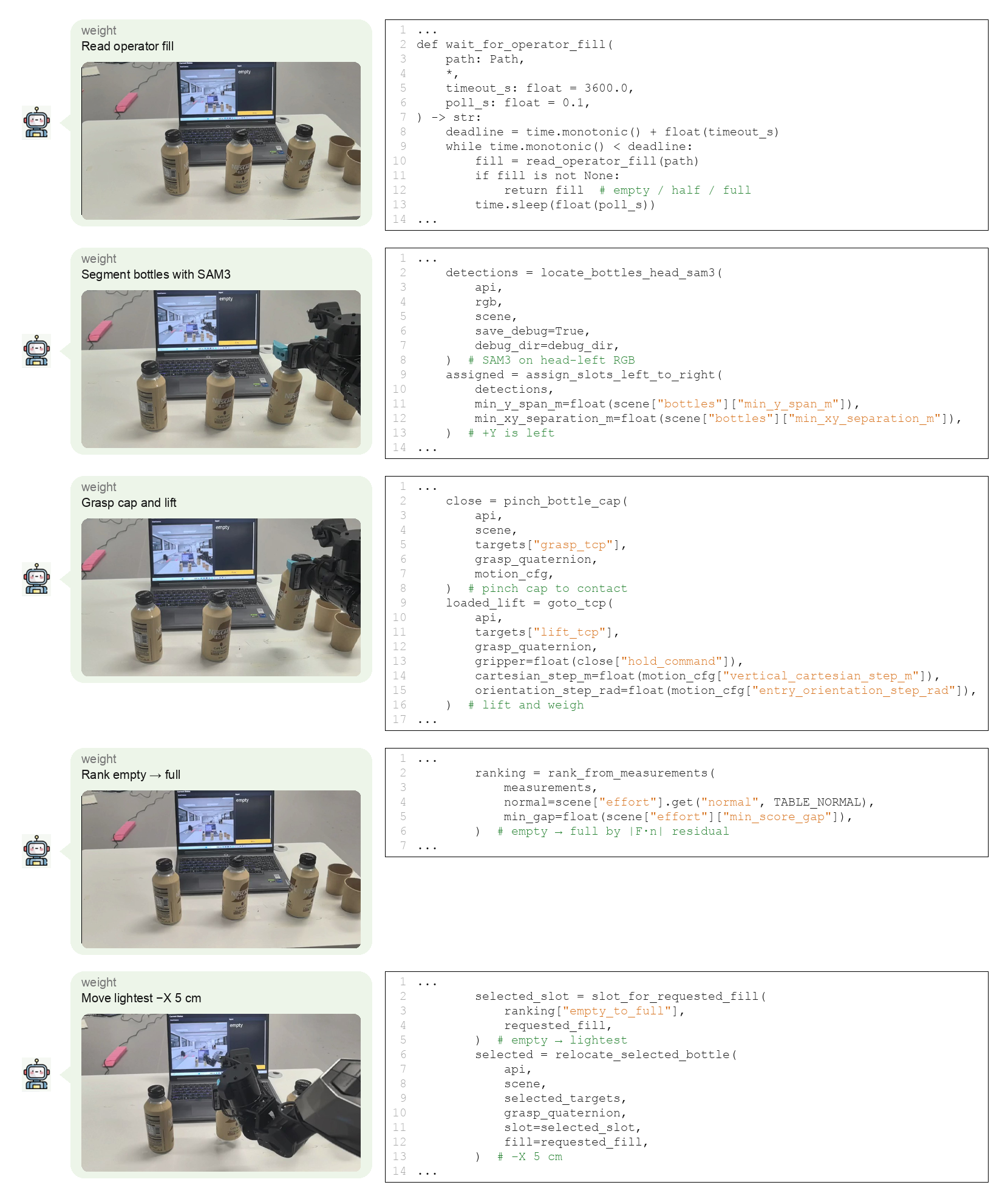}
    \caption{We validate the weighing capability of META-Robot on a real-world Galaxea R1 Lite using three visually identical plastic bottles whose fill levels---empty, half, and full---cannot be recovered from RGB appearance. META-Robot uses its proprioceptive signal to realize this task with high accuracy.}
    \label{fig:Weight comparison}
\end{figure}

The operator specifies the desired fill in a Current Status window; Enter submits empty, half, or full and starts motion. No bottle is labeled, and the prompt provides no spatial hint for which slot is lightest.\par
By its proprioceptive signal orchestration mechanism, META-Robot decomposes the task into perceive, weigh, rank, and relocate. SAM3 segments the standing bottles from the head-left camera; slots are assigned left-to-right in the robot base, while the right arm weighs them from right to left.\par
The key is the signal of joint torques. Each weighing cycle pinches the cap, lifts, and estimates table-normal force from joint effort via $\tau \approx J^\mathrm{T} F$. The meta-robot then computes a load score: $|F\cdot n|_{\text{loaded}} - |F\cdot n|_{\text{empty}}$. After three measurements, META-Robot generates three functions: it first sorts the bottles from empty to full, then maps the operator token onto that ranking from empty to the lightest, half to the middle, full to the heaviest, after that it shifts that bottle 5 cm along robot-reverse, which corresponds to the selection order.\par

\clearpage
\begin{figure}[H]
\centering
    \includegraphics[width=.95\linewidth]{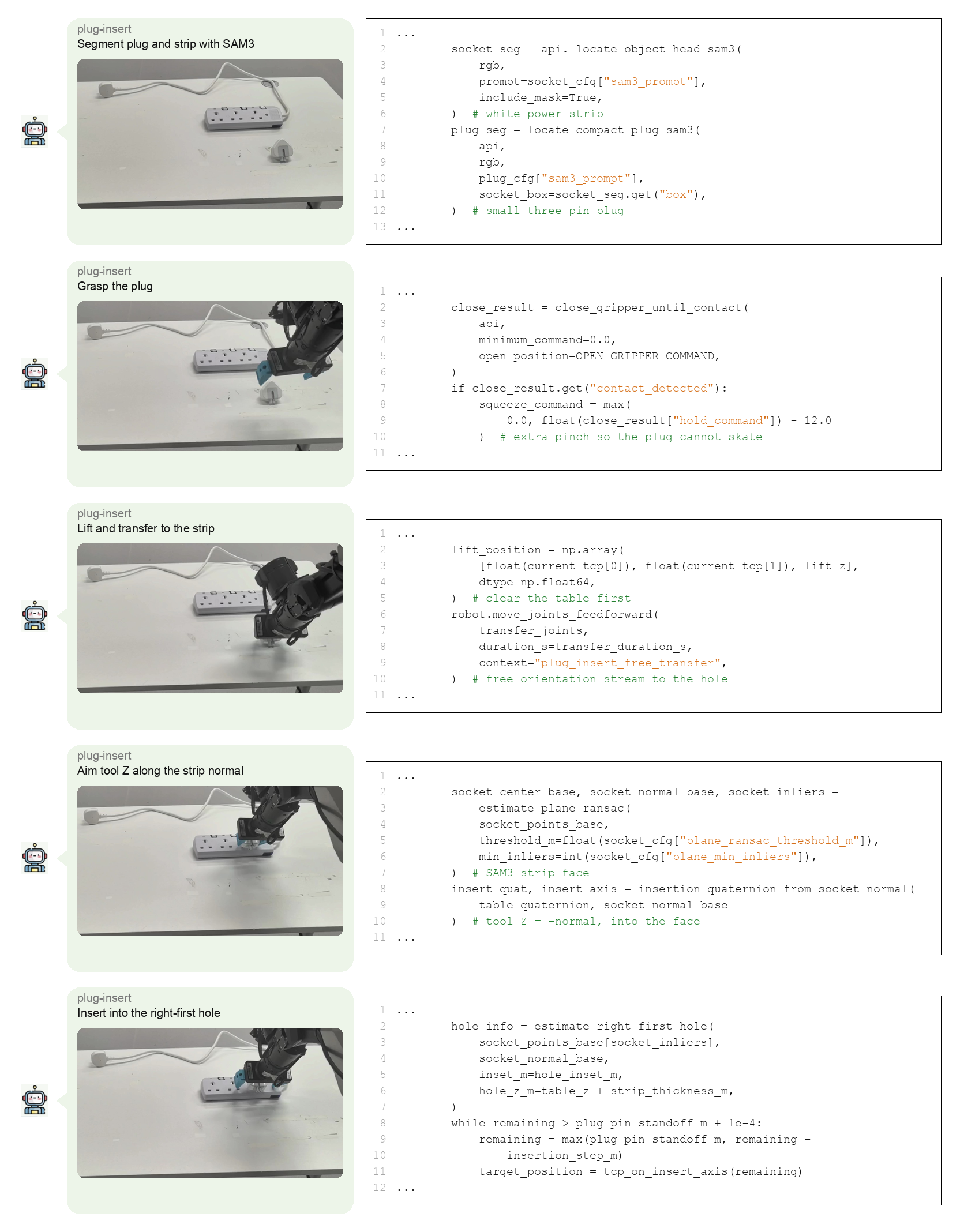}
    \caption{We validate a precision plug-insertion task of META-Robot on a real-world Galaxea R1 Lite. A small three-pin plug and a white power strip are placed at randomized poses on the table. This task shows the use of proprioceptive signal in precise task.}
    \label{fig:Plug insertion}
\end{figure}

META-Robot decomposes the task into perceive, grasp, transfer, align, and insert. SAM3 first segments the strip and the compact plug. The outward face normal comes from the plane of the detected power strip. The target is the robot-right first hole, inset from the right end of the strip.\par
After grasping the plug at its angle, the arm lifts, then streams with free orientation to the hole. Near the target hole of the power strip, meta-robot adjusts the angle of the gripper to make sure the plug is perpendicular to the surface normal of the power strip in the opposite direction. The angle alignment is the most difficult part. META-Robot aligns the plug's pin with the hole based on the initially identified position by SAM3. The insertion then advances along that axis until the pin standoff, opens the gripper, and returns home.\par
\clearpage
\subsection{subtask analysis of zero-shot transfer}
\label{app:subtask analysis of zero-shot transfer}
\begin{figure}[!htbp]
    \centering
    \includegraphics[width=.9\linewidth]{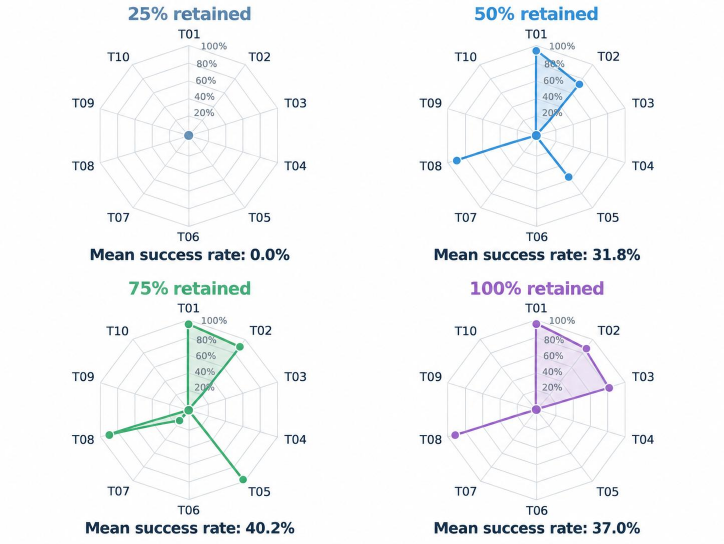}
    \caption{Zero-shot transfer to LIBERO-Pro Long tasks: we evaluate ten representative manipulation tasks under 25\%, 50\%, 75\%, and 100\% information retention. The tasks include T01 soup-and-sauce placement, T02 cheese-and-butter placement, T03 stove/moka-pot interaction, T04 bowl-in-drawer placement, T05 mug-to-plate placement, T06 book-to-caddy placement, T07 mug-and-pudding arrangement, T08 soup-and-cheese placement, T09 moka-pot placement, and T10 mug-in-microwave placement. Each radar plot reports per-task success rates, with the mean success rate shown below.}
    \label{fig:appendix-1}
\end{figure}
\FloatBarrier
\subsection{API choice}
\label{app:API choice}
\begin{figure}[!htbp]
    \centering
    \includegraphics[width=1\linewidth]{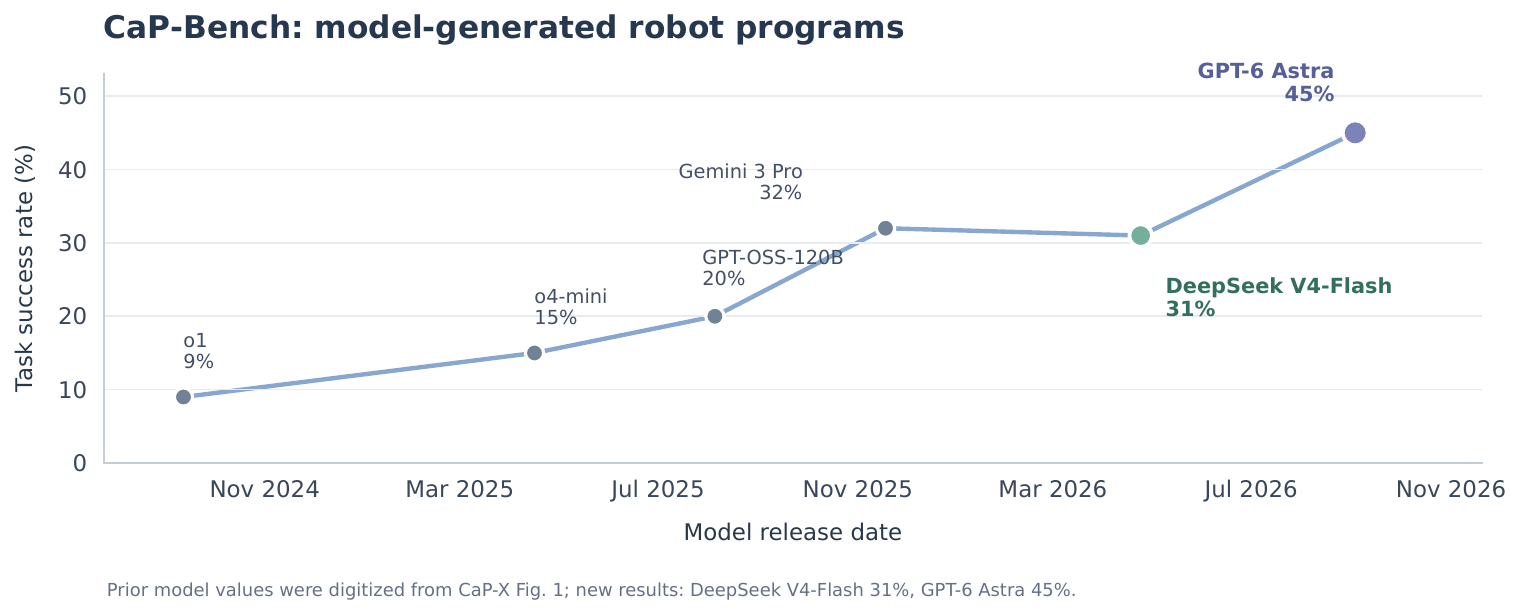}
    \caption{We extend the base model comparison shown in CaP-X \cite{fu2026cap}with DeepSeek-V4-Flash and GPT-6 Astra, reporting task success rates of 31\% and 45\%, respectively.}
    \label{fig:appendix-2}
\end{figure}

GPT-6 Astra reaches 45\%, above the highest historical point shown in the figure, Gemini 3 Pro at about 32\%. DeepSeek-V4-Flash reaches 31\%, close to that earlier result. The added evaluations therefore show a clear gain for GPT-6 Astra, while DeepSeek-V4-Flash performs at a similar level to the strongest model highlighted in the original plot.

\FloatBarrier
\clearpage
\subsection{Skill library \& Predicate library Algorithm}
\label{app:Predicae &Skill library}
\begin{algorithm}[H]
\caption{Coupled Skill and Action-Level Predicate Library Consolidation}
\label{alg:skill_predicate_consolidation}
\begin{algorithmic}[1]
\Require Skill forest $\mathcal{L}_{S}$,
action-level predicate library $\mathcal{P}$,
associations $\Gamma$, candidate repair $\rho$,
and execution evidence $\mathcal{D}$
\Ensure Updated $\mathcal{L}_{S}$, $\mathcal{P}$, and $\Gamma$

\If{$\neg\operatorname{Validated}(\rho,\mathcal{D})$}
    \State \Return $(\mathcal{L}_{S},\mathcal{P},\Gamma)$
\EndIf

\State $(\hat{s},\widehat{\mathcal{P}})
    \gets \operatorname{Distill}(\rho,\mathcal{D})$
\State $\mathcal{V}_{\mathrm{rel}}
    \gets \operatorname{RetrieveRelated}
    (\hat{s},\mathcal{L}_{S})$
\State $(\mathcal{L}'_{S},v)
    \gets \operatorname{UpdateSkillTree}
    (\hat{s},\mathcal{V}_{\mathrm{rel}},\mathcal{L}_{S})$

\If{$\operatorname{NeedsValidation}(\mathcal{L}'_{S},v)
    \land
    \neg\operatorname{ValidateSkillUpdate}
    (\mathcal{L}'_{S},v,\mathcal{D})$}
    \State \Return $(\mathcal{L}_{S},\mathcal{P},\Gamma)$
\EndIf

\State $\mathcal{P}'\gets\mathcal{P}$
\State $\Gamma'\gets\Gamma$

\For{$\hat{\phi}\in\widehat{\mathcal{P}}$}
    \State $\phi\gets
        \operatorname{FindEquivalentPredicate}
        (\hat{\phi},\mathcal{P}')$

    \If{$\phi=\emptyset$}
        \If{$\neg\operatorname{ValidatePredicate}
            (\hat{\phi},\mathcal{D})$}
            \State \textbf{continue}
        \EndIf

        \State $\phi\gets\hat{\phi}$
        \State $\mathcal{P}'\gets
            \mathcal{P}'\cup\{\phi\}$
    \EndIf

    \State $\Gamma'(v)\gets\Gamma'(v)\cup
        \left\{
        \left(
        \phi,
        \operatorname{EvalMeta}(\hat{\phi},\mathcal{D})
        \right)
        \right\}$
\EndFor

\State $(\mathcal{P}',\Gamma')
    \gets\operatorname{ReindexPredicates}
    (\mathcal{P}',\Gamma',\mathcal{L}'_{S})$

\State \Return $(\mathcal{L}'_{S},\mathcal{P}',\Gamma')$
\end{algorithmic}
\end{algorithm}
\end{document}